\documentclass{article}

\usepackage[preprint]{neurips_2026}

\usepackage[utf8]{inputenc}
\usepackage[T1]{fontenc}
\usepackage{hyperref}
\usepackage{url}
\usepackage{booktabs}
\usepackage{amsfonts}
\usepackage{amsmath}
\usepackage{amssymb}
\usepackage{nicefrac}
\usepackage{microtype}
\usepackage{xcolor}
\usepackage{graphicx}
\usepackage{subcaption}
\usepackage{longtable}

\newif\ifdraft
\draftfalse

\newcommand{\todo}[1]{\ifdraft\textcolor{red}{[TODO: #1]}\fi}

\newcommand{\method}{\textit{TFM-Retouche}}        
\newcommand{\specmethod}{\textit{TabICLv2-Retouche}} 
\newcommand{\TabICL}{TabICL}
\newcommand{\TabPFN}{TabPFN}

\newcommand{\adapter}{g_\phi}
\newcommand{\frozenTFM}{f_\theta}

\title{\method{}: A Lightweight Input-Space Adapter for Tabular Foundation Models}

\author{%
  Duong Nguyen \quad Mohammed Jawhar \quad Nicolas Chesneau \\
  Ekimetrics \\
}

\begin{document}

\maketitle

\begin{abstract}
Tabular foundation models (TFMs), such as TabPFN-2.6, TabICLv2, ConTextTab, Mitra, LimiX, and TabDPT, achieve strong zero-shot performance through in-context learning, but their inductive biases remain fixed at inference time. Adapting a pretrained TFM to a specific dataset or task typically requires either full fine-tuning, which is computationally expensive, or parameter-efficient tuning methods (PEFT) such as LoRA, which must be tailored to the internal architecture of each TFM. Furthermore, the evidence on whether weight-space fine-tuning improves accuracy or calibration is mixed \citep{tanna_exploring_2026,rubachev_finetuning_2025}.
We introduce \method{}, a lightweight \textit{input-space} residual adapter that is architecture-agnostic by design with respect to the frozen TFM backbone. \method{} learns a small residual correction in the input space to align the input data with the inductive biases of the pretrained model. The adapter is trained end-to-end through the frozen TFM,  with a post-training identity guard that falls back to the unmodified TFM whenever adaptation does not help on held-out validation. On TabArena-Lite \citep{erickson_tabarena_2025} (51 datasets spanning binary classification, multiclass classification, and regression), \specmethod{}---the framework instantiated on \TabICL{}v2---is the top-ranked method on the leaderboard with light per-task tuning and ensembling, lifting aggregate Elo by $+56$ over the frozen \TabICL{}v2 base and sitting on the Pareto front of predictive quality versus both training and inference time.


\begin{figure}[h]
\centering
\includegraphics[width=1.0\linewidth]{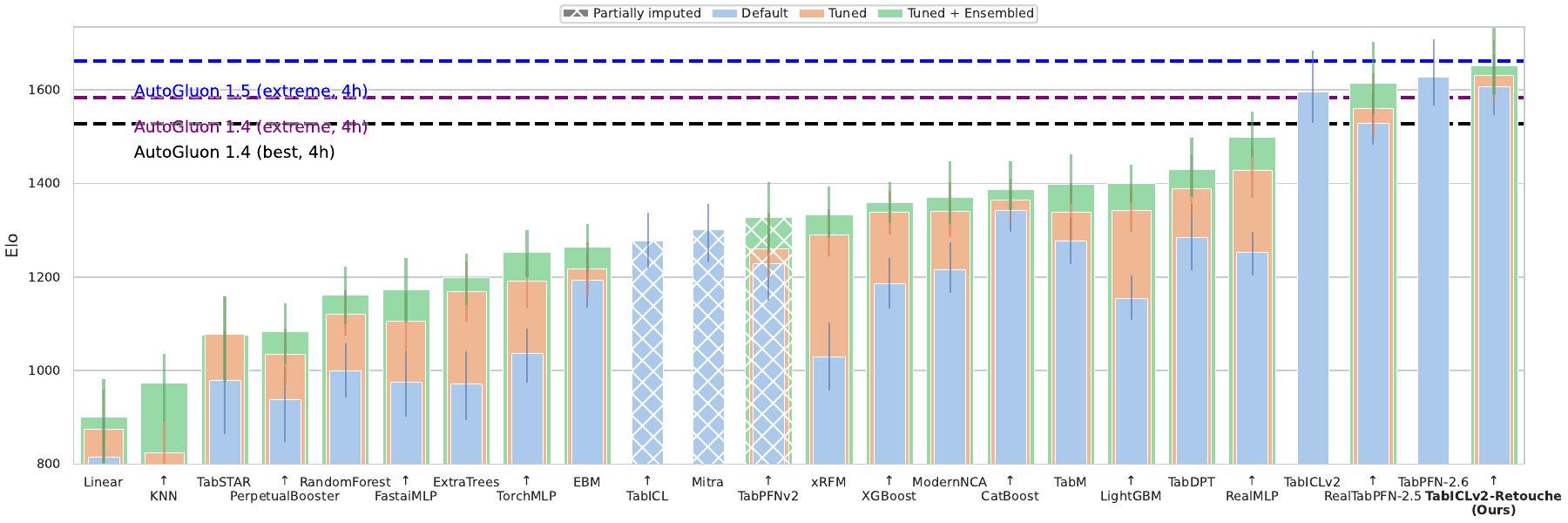}
\caption{TabArena-Lite leaderboard. \specmethod{} (T+E)---our framework instantiated on frozen \TabICL{}v2---is the top-ranked method, achieved under a $20\times$ smaller per-dataset HPO budget than every tuned peer ($10$ random configurations vs.\ TabArena's standard $200$).}
\label{fig:tuning-impact}
\end{figure}

\end{abstract}

\section{Introduction}
\label{sec:intro}

Tabular data is central to many practical machine-learning applications, from healthcare and finance to enterprise analytics~\citep{borisov_deep_2024, shwartz-ziv_tabular_2022, fatima_survey_2017, dastile_statistical_2020}. The field was long dominated by gradient-boosted decision trees such as XGBoost~\citep{chen_xgboost_2016}, LightGBM~\citep{ke_lightgbm_2017}, and CatBoost~\citep{prokhorenkova_catboost_2019}, and recent deep architectures such as RealMLP~\citep{holzmuller_better_2025} and TabM~\citep{gorishniy_tabm_2025} have closed, and in places overtaken, the long-standing gap with well-tuned trees. Tabular foundation models (TFMs)---\TabPFN{}~\citep{hollmann_tabpfn_2023}, \TabPFN{}v2~\citep{hollmann_accurate_2025}, \TabPFN{}-2.5~\citep{grinsztajn_tabpfn-25_2026}, \TabPFN{}-2.6~\citep{grinsztajn_tabpfn-25_2025}, Mitra~\citep{zhang_mitra_2025}, LimiX~\citep{zhang_limix_2025}, ConTextTab~\citep{spinaci_contexttab_2025}, TabDPT~\citep{ma_tabdpt_2026}, \TabICL{}~\citep{qu_tabicl_2025}, and \TabICL{}v2~\citep{qu_tabiclv2_2026}---have reshaped this picture: by pretraining at scale on synthetic data drawn from explicit prior generators, they encode a transferable inductive bias for tabular prediction that often rivals or surpasses tuned tree ensembles on small-to-medium problems~\citep{erickson_tabarena_2025}.

TFMs predict zero-shot through In-Context Learning (ICL). A labeled context is appended to the query and the prediction emerges from a single forward pass through frozen weights, with no task-specific training~\citep{breugel_why_2024}. The conditioning step does substantial work: the in-context examples pull the prior carried by the pretrained model toward the real data of the downstream task. Yet a residual gap remains: per-task structure such as domain feature semantics, monotonicities, and dominant interactions is not always captured by what the synthetic generators emphasize, and zero-shot ICL forgoes accuracy and calibration gains that targeted adaptation can capture.

Existing approaches to that adaptation have predominantly targeted the model's weights. However, full fine-tuning is computationally expensive, and parameter-efficient fine-tuning (PEFT) methods such as LoRA~\citep{hu_lora_2021} are lighter but tied to the backbone's internal architecture, requiring per-model choices about where and how to insert trainable modules. Beyond these practical costs, recent empirical studies have reported mixed effects on accuracy and calibration when weight-space adaptation is applied to TFMs~\citep{rubachev_finetuning_2025,tanna_exploring_2026}. An interpretation is that the synthetic prior carried by the pretrained weights is fragile under further gradient updates, since the inductive bias that makes a TFM strong out-of-the-box can be eroded by even modest weight-space changes, leaving these methods with a narrow operating margin between under- and over-adaptation.

We instead adapt the \emph{input}, not the weights. Prior TFM input-space adaptation (BETA~\citep{liu_tabpfn_2025}, discussed in Section~\ref{sec:related:adapt}), replaces the original features with a learned encoding. That design is shaped by first-generation \TabPFN{}'s capacity constraints. With recent TFMs such as \TabPFN{}-2.5, \TabPFN{}-2.6, and \TabICL{}v2 having largely lifted those capacity constraints, the bottleneck is no longer capacity but \emph{alignment}: the frozen model's biases come from a generic synthetic prior that need not match the structure of any specific downstream task.

Our framework, \emph{TFM-Retouche} (\textbf{T}abular \textbf{F}oundation \textbf{M}odel \textbf{Re}sidual \textbf{touch}ed \textbf{e}xtension), targets alignment by \emph{nudging} the input rather than replacing it. A near-identity residual adapter applies a small, learned correction to the original input to align it with the inductive biases the frozen model expects, preserving feature dimensionality, treating the pretrained model as a differentiable but otherwise unmodified black box, and requiring no internal modifications. To our knowledge, this residual input-space adaptation formulation has not previously been studied for TFMs.

Our contributions are as follows:
\begin{enumerate}
\item The \method{} framework: We introduce a lightweight, configurable input-space residual adapter for frozen tabular foundation models. The adapter is dimension-preserving, architecture-agnostic with respect to the backbone TFM, and supports both classification and regression objectives. Its inner block is modular: we instantiate it as either a DCNv2 cross network~\citep{wang_dcn_2021} or a residual-bottleneck MLP block.

\item End-to-end training through a frozen TFM with post-training identity guard: We train the adapter end-to-end by backpropagating loss through the frozen TFM forward pass, leaving every pretrained weight unchanged. A per-fit identity guard scores the adapter path against the unmodified TFM and routes around the adapter at inference whenever adaptation does not improve, bounding the risk of adaptation.

\item Evaluation on TabArena-Lite: We provide a thorough evaluation \specmethod{}---the framework instantiated on \TabICL{}v2---on TabArena-Lite (51 datasets spanning binary classification, multiclass classification, and regression), where it is the top-ranked method on the leaderboard with light per-task tuning and ensembling. Ablation studies are presented in Appendix~\ref{app:ablations}. We also include in Appendix~\ref{app:h2h} head-to-head comparisons against the three contemporary adaptation paradigms applicable to a frozen \TabICL{}v2 backbone: input-space encoding via BETA~\citep{liu_tabpfn_2025} on the classification subset, LoRA-based PEFT, and full supervised fine-tuning. Appendix~\ref{app:talent} extends the evaluation to the TALENT benchmark~\citep{liu_talent_2024}.

\item \textbf{Open-source release.} We will release \method{} as an open-source library, with different configurable settings to help practitioners adapt pretrained TFMs to their specific tasks.
\end{enumerate}

\section{Related Work}
\label{sec:related}

\subsection{Tabular Foundation Models}

Tabular foundation models predict zero-shot through In-Context Learning (ICL): a labeled context is appended to the query and the prediction emerges from a single forward pass through frozen weights. This paradigm was introduced by \TabPFN{}~\citep{hollmann_tabpfn_2023}---a transformer pretrained on synthetic data to perform approximate Bayesian inference over a tabular prior, delivering strong zero-shot performance without task-specific gradient updates. 
We focus throughout on transformer-based ICL TFMs that follow this template.  Alternative approaches, such as Carte~\citep{kim_carte_2024} which pretrains a graph neural network on a real-data corpus, fall outside the scope of the present study.

Subsequent work has extended this paradigm along three axes. \emph{Model capacity}: alternating row- and column-wise attention in \TabPFN{}v2~\citep{hollmann_accurate_2025}, further sample/feature expansion in \TabPFN{}-2.5~\citep{grinsztajn_tabpfn-25_2026} and \TabPFN{}-2.6~\citep{grinsztajn_tabpfn-25_2025}, column-then-row attention in \TabICL{}~\citep{qu_tabicl_2025} with further architectural improvements in \TabICL{}v2~\citep{qu_tabiclv2_2026}, and multi-scale processing with block-sparse attention in Orion-MSP~\citep{bouadi_orion-msp_2025}. \emph{Pretraining prior quality}: Mitra~\citep{zhang_mitra_2025} trains on a curated mixture of synthetic priors; \TabICL{}v2~\citep{qu_tabiclv2_2026} introduces a synthetic data generation engine designed for high pretraining diversity. \emph{Mechanism}: LimiX~\citep{zhang_limix_2025} extends ICL to joint-distribution modeling over masked inputs, enabling query-based conditional prediction; ConTextTab~\citep{spinaci_contexttab_2025} adds specialized embeddings for different data modalities; TabDPT~\citep{ma_tabdpt_2026} uses retrieval-based pretraining on real data; and TabSTAR~\citep{arazi_tabstar_2025} departs from the synthetic-prior ICL template entirely, combining multi-task supervised pretraining on real tabular datasets over a pretrained text encoder backbone with per-dataset LoRA fine-tuning conditioned on verbalized target tokens.

Across all three axes, the open question for practitioners is alignment: a TFM trained on a generic synthetic prior is not specialized to any downstream task.

\subsection{Adapting TFMs to downstream tasks}
\label{sec:related:adapt}

Once a TFM has been pretrained, the question becomes how to specialize its behavior to a particular downstream dataset. Existing work splits along two axes: \emph{weight-space} adaptation, which updates the pretrained parameters of the backbone, and \emph{input-space} adaptation, which leaves the backbone untouched and instead transforms the features that flow into it.

\textbf{Weight-space adaptation (fine-tuning)}: Weight-space adaptation updates backbone parameters either fully or through a parameter-efficient interface such as LoRA~\citep{hu_lora_2021}. Two recent studies provide the most relevant empirical evidence. \citet{rubachev_finetuning_2025} evaluates fine-tuning strategies for \TabPFN{}v2 on i.i.d.\ academic benchmarks and finds improvements over zero-shot, but reduced stability under temporal shift and on feature-rich datasets. \citet{tanna_exploring_2026} uses the TabTune library~\citep{tanna_tabtune_2025} for a cross-model comparison across TALENT, OpenML-CC18, and TabZilla and finds that gains are highly model- and data-dependent: meta-learning and PEFT yield only moderate improvements under specific conditions, and full SFT often degrades accuracy and calibration. Neither study evaluates the latest TFM generation (\TabICL{}v2~\citep{qu_tabiclv2_2026}, \TabPFN{}-2.5, or \TabPFN{}-2.6~\citep{grinsztajn_tabpfn-25_2025}). Hence, whether their conclusions generalize to this latest generation remains open.

Beyond these empirical uncertainties, practical costs accrue regardless of backbone: full fine-tuning demands backpropagation and optimizer state for every backbone parameter, while PEFT methods such as LoRA are tied to the backbone's module layout, since the choice of which linear projections to wrap and at what rank must be made per architecture. 

\textbf{Input-space adaptation}: The closest prior work is BETA~\citep{liu_tabpfn_2025}: a two-layer MLP encoder (hidden and output dimension 100, ReLU and periodic activations) trained end-to-end before a frozen \TabPFN{}, with 16 bootstrap-sampled encoders bagged at inference. The fixed-width bottleneck reflects the narrow context capacity of first-generation \TabPFN{}, where compressing a high-dimensional input was practically necessary. With newer TFMs such as \TabICL{}v2, \TabPFN{}-2.5, and \TabPFN{}-2.6 handling much larger inputs natively, the central adaptation question shifts from capacity to alignment. \method{} targets alignment through a different interface and differs from BETA in three respects: (i) \emph{dimension-preserving} rather than fixed-width bottleneck; (ii) it \emph{nudges} the input by a small learned correction, whereas BETA \emph{replaces} the input with a freshly learned encoding; and (iii) applicable to both classification and regression, whereas BETA is classification-only.

\section{Method}
\label{sec:method}

\subsection{Problem formulation}

Let $\frozenTFM$ denote a pretrained tabular foundation model with frozen parameters $\theta$, and let $D^{raw} = \{(x^{raw}_i, y^{raw}_i)\}_{i=1}^N$ be a downstream dataset and $D = \{(x_i, y_i)\}_{i=1}^N$ the preprocessed data. $x_i \in \mathbb{R}^d$ is feature vector and $y_i$ is the target. We seek a lightweight, parameter-efficient adapter $\adapter : \mathbb{R}^d \to \mathbb{R}^d$, parameterized by $\phi$ with $|\phi| \ll |\theta|$ such that $\frozenTFM(\adapter(X))$ outperforms $\frozenTFM(X^{raw})$.

Three design choices distinguish $\adapter$ from prior adaptation paradigms (Section~\ref{sec:related:adapt}). It is (i) \emph{dimension-preserving} (in contrast to BETA's fixed-width encoder~\citep{liu_tabpfn_2025}); (ii) \emph{initialized near identity} 
; and (iii) acts \emph{only at the input boundary} through a differentiable forward pass of $\frozenTFM$---a more general interface than weight-space PEFT methods such as LoRA, which must be tailored per backbone.

\begin{figure}
    \centering
    \includegraphics[width=1.0\linewidth]{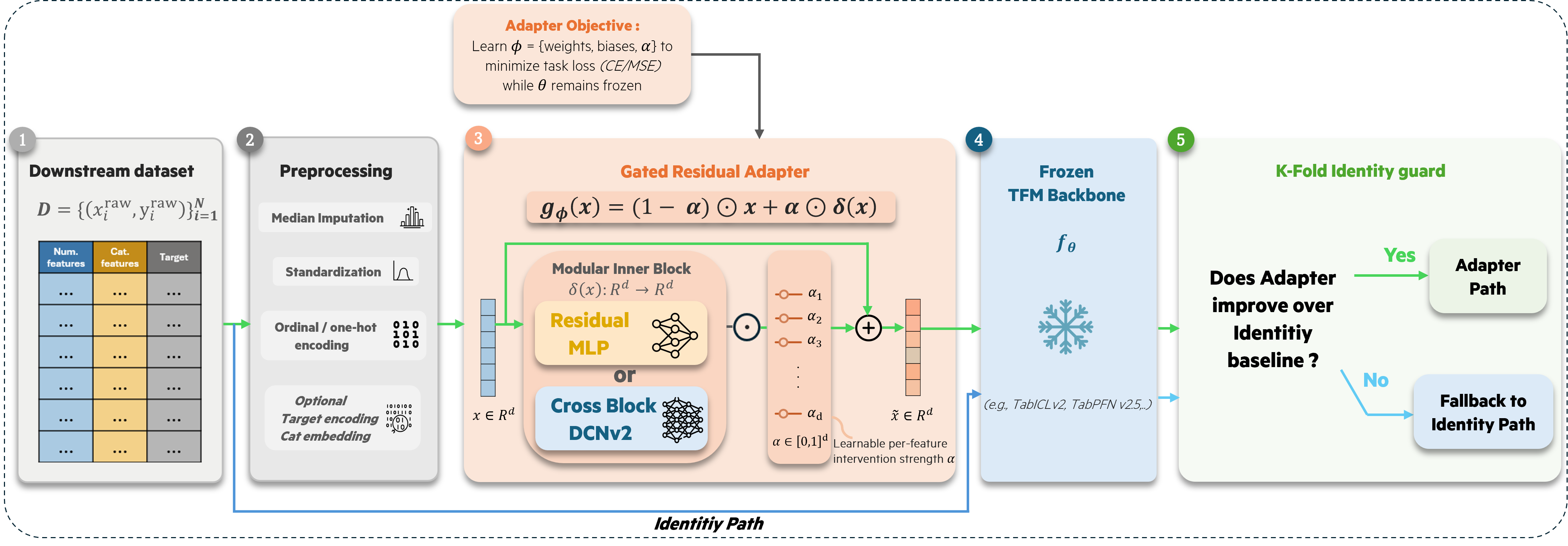}
    \caption{\textbf{Overview of \method{}.} A preprocessed input $x$ is passed through a gated residual adapter $\adapter(x) = (1-\alpha) \odot x + \alpha \odot \delta(x)$ before entering a frozen TFM. Only the adapter parameters $\phi$ are trained end-to-end through the frozen backbone $f_{\theta}$. After training, an identity guard falls back to the unmodified TFM when adaptation does not improve held-out validation performance.}
    \label{fig:architecture}
\end{figure}

Figure~\ref{fig:architecture} summarizes the framework. We describe each component in detail below.

\subsection{The adapter}
\label{sec:method:adapter}

The core design is a gated adapter with a learnable strength parameter:
\begin{equation}
\adapter(x) \;=\; (1-\alpha) \odot x \;+\; \alpha \odot \delta(x),
\label{eq:adapter}
\end{equation}
where $\alpha$ is a learnable parameter, $\delta : \mathbb{R}^d \to \mathbb{R}^d$ is a trainable transformation, and $\odot$ denotes the Hadamard product. This construction preserves the input dimensionality while allowing the adapter to apply a task-specific correction to the original feature representation. Equation~\ref{eq:adapter} can equivalently be written as a pure additive residual on $x$, $\adapter(x) = x + \alpha \odot \big(\delta(x) - x\big)$, so the convex-blend and pure-residual viewpoints coincide. Appendix~\ref{app:residual-form} gives the derivation and shows why the equivalence holds in closed form for both inner blocks of Section~\ref{sec:method:blocks}.

As default, we treat $\alpha$ as a vector $\alpha \in \mathbb{R}^d$: so that each feature dimension has its own learnable intervention strength. This lets the adapter correct some channels aggressively while leaving others close to identity. As a lower-capacity alternative, we also consider a scalar gate $\alpha \in \mathbb{R}$, that intervenes uniformly across all features. With $d$-fold fewer free gating parameters than the per-channel form, the scalar gate may act as a regularizer when the per-channel gate is prone to overfitting, although a controlled comparison is left to future work. Both shapes broadcast over $(N, d)$ inputs identically and differ only in capacity.

We initialize $\alpha$ to a small value $\alpha_0 = 0.02$, so that $\adapter(x) \approx x$ at the start of training and identity is recovered exactly when $\alpha = 0$ (Appendix~\ref{app:residual-form}). The motivation is intrinsic to the problem: the frozen backbone already carries a strong default performance, so we want the TFM to see inputs essentially indistinguishable from those it learned on at the start of training, and depart from this near-identity regime to amplify or suppress the correction only as the loss signal opens specific channels. This small-initialization choice is similar in spirit to ReZero~\citep{bachlechner_rezero_2020} and LayerScale~\citep{touvron_going_2021}.

\subsection{Inner block: two variants}
\label{sec:method:blocks}

The inner-block transformation $\delta : \mathbb{R}^d \to \mathbb{R}^d$ is implemented as a stack of $L$ residual layers, where the depth $L$ is a hyperparameter. We study two variants that induce different biases while remaining interchangeable within the outer adapter of Equation~\ref{eq:adapter}.

\paragraph{Cross block (DCNv2).}
\begin{equation}
\delta_{\text{cross}}(x) \;=\; x_L, \qquad x_{l+1} \;=\; x_0 \odot (W_l x_l + b_l) + x_l, \quad l=0,\dots,L-1.
\label{eq:cross-block}
\end{equation}
where $W_l = U_l V_l^\top$ with $U_l, V_l \in \mathbb{R}^{d \times h}$, $h = \lfloor r d \rfloor$, and $r$ is the low-rank ratio, following \citet{wang_dcn_2021}. When the inner block is low-rank, an optional activation $\sigma$ may be inserted between $U_l$ and $V_l$, i.e.\ $W_l x_l$ is replaced by $V_l\,\sigma(U_l\, x_l)$~\citet{wang_dcn_2021}; $\sigma$ is a hyperparameter with the identity as default (see Appendix~\ref{app:hpo}). Each layer modulates the current representation $x_l$ channel-wise by the original input $x_0$. Stacking $L$ such layers yields explicit polynomial feature interactions of degree up to $L+1$~\citet{wang_dcn_2021}. The Hadamard product with the original input $x_0$ imposes an \textit{explicit} multiplicative structure and anchors each output channel to its corresponding input channel. The fitted weights are directly inspectable as pairwise feature interactions (Appendix~\ref{app:weight-inspection}).

\paragraph{Residual MLP block.}

\begin{equation}
\delta_{\text{MLP}}(x) \;=\; x_L, \qquad x_{l+1} \;=\; x_l + W_l^{2} \, \sigma\!\left(W_l^{1} x_l + b_l^{1}\right) + b_l^{2}, \quad l=0,\dots,L-1,
\label{eq:mlp-block}
\end{equation}
where $W_l^{1} \in \mathbb{R}^{h \times d}$, $W_l^{2} \in \mathbb{R}^{d \times h}$, and the hidden width is $h = \max(h_{\min}, \lfloor r d \rfloor)$ for expansion ratio $r$ and a minimum hidden width $h_{\min}$ (default $2$). Each layer first projects the input to width $h$, applies activation $\sigma$ (default ReLU), and projects back to $d$, with an optional BatchNorm at the layer output. The per-layer residual connection initializes each block close to the identity and stabilizes optimization. Structurally, this block implements a dense nonlinear transformation in which every output channel is a learned mixture of all input features, with no explicit multiplicative structure or per-channel anchoring.

The two blocks share the same outer interface, depth $L$, low-rank/expansion ratio $r$, BatchNorm option, and initialization scheme, and are matched at $O(2dh)$ trainable weights per layer at fixed $r$, with biases and the optional BatchNorm contributing an additional $O(d)$ parameters per layer. In the TabArena-Lite experiments (Section~\ref{sec:exp:main}), we use the DCNv2 cross block as the default and treat the residual-MLP variant as a targeted ablation (Appendix~\ref{app:ablations}).

\subsection{End-to-end training}
\label{sec:method:training}

We train the adapter parameters end-to-end while keeping the backbone TFM fully frozen. During each epoch, the training data are randomly partitioned into a context set $\mathcal{C}$ and a query set $\mathcal{Q}$, matching the in-context-learning setting of the backbone. 

We optimize the adapter with AdamW \citep{loshchilov_decoupled_2019}, using separate parameter groups for weight matrices, biases, and the gate $\alpha$. Weight decay is applied only to the matrix parameters, while biases and $\alpha$ are left unregularized. We further assign $\alpha$ a larger learning rate than the remaining parameters (a factor of 3, in our TabArena instantiation). As an alternative to using AdamW on the matrix parameters, we also consider Muon~\citep{jordan_muon_2024}, while keeping AdamW for biases and $\alpha$, since Muon is defined only for two-dimensional parameters. We use cross-entropy with optional label smoothing for classification and mean squared error for regression. The default learning-rate schedule is a multi-cycle log-spaced cosine schedule \citep{holzmuller_better_2025}, with single-cycle cosine and constant alternatives included in the hyperparameter sweep.

\subsection{Identity guard}
\label{sec:method:guard}

Once training is complete, the identity guard decides, per fit, whether the trained adapter is actually used at inference or bypassed in favor of the unmodified TFM.

The guard takes an held out validation set and scores two estimators on it: the adapter path $\frozenTFM(\adapter(X))$ vs. the TFM baseline that consumes the raw input $\frozenTFM(X^{raw})$. Scoring uses the deployment metric (for TabArena-Lite: $1-\mathrm{AUC}$ for binary classification, log loss for multiclass, and mean squared error for regression) rather than the training loss, so that the guard's gating decision reflects inference-time performance directly. If the adapter path does not improve over the plain baseline by a small tolerance (default $0.5\%$), the outer estimator routes around the adapter entirely at inference and the prediction is identical to the unmodified TFM. 

We report identity-fallback rates of the experiments presented in Section~\ref{sec:exp} in Appendix~\ref{app:analysis}.


\subsection{Implementation details}
\label{sec:method:impl}

The remaining implementation choices (preprocessing of numeric and categorical features and the alternative encoders considered, the input-dimension cap with its trainable low-rank projection for high-dimensional datasets, batch-size heuristic, per-epoch train/validation reshuffling, and the automatic-precision policy of the frozen TFM forward, etc.) are deferred to Appendix~\ref{app:impl}. They affect efficiency and numerical conditioning but leave the adapter formulation of Equation~\ref{eq:adapter} unchanged.

\section{Experiments}
\label{sec:exp}

\subsection{Setup}
\label{sec:exp:setup}

\paragraph{Benchmark.}
We evaluate \method{} on TabArena-Lite~\citep{erickson_tabarena_2025}, following the standard TabArena protocol for data splits, model comparison, and aggregate reporting.
TabArena-Lite is a curated 51-dataset covering binary classification, multiclass classification, and regression. The datasets range from 748 to 150{,}000 samples and from 5 to 1{,}777 features, providing a broad mix of small, medium, and moderately high-dimensional tabular problems. We adopt the standard AutoGluon/TabArena evaluation protocol based on 8-fold bagged cross-validation throughout. Within each fold, AutoGluon holds out a per-fold validation split that we reuse as the external validation set for the identity guard of Section~\ref{sec:method:guard}.


\paragraph{Instantiation of \method{}.}
We denote by \specmethod{} the instantiation of \method{} using frozen \TabICL{}v2~\citep{qu_tabiclv2_2026} as the backbone TFM. We choose \TabICL{}v2 because it is one of the strongest open tabular foundation models currently available and exposes a forward pass that is convenient for our end-to-end adaptation setting. Another strong TFM is \TabPFN{}-2.6. However, we can not use this model because of the license. The adaptation framework itself is backbone-agnostic in design, but in this paper we benchmark only the \TabICL{}v2 instantiation. Extending the same interface to other openly available TFMs is left to future work (Section~\ref{sec:limits}).

\paragraph{Baselines.}
Our primary baseline is the frozen \TabICL{}v2 backbone without adaptation, since this isolates the contribution of the adapter from that of the underlying TFM. To place \specmethod{} in the broader tabular-learning landscape, we also compare against strong leaderboard baselines from TabArena-Lite, including RealTabPFN-2.5 and \TabPFN{}-2.6~\citep{grinsztajn_tabpfn-25_2026}, the standard tabular suite comprising XGBoost~\citep{chen_xgboost_2016}, LightGBM~\citep{ke_lightgbm_2017}, CatBoost~\citep{prokhorenkova_catboost_2019}, RealMLP~\citep{holzmuller_better_2025}, TabM~\citep{gorishniy_tabm_2025}, TabDPT~\citep{ma_tabdpt_2026}, ModernNCA~\citep{ye_revisiting_2025}, EBM~\citep{nori_interpretml_2019}, as well as additional tree-based and linear baselines (full list in Appendix~\ref{app:leaderboard}). All baseline results are taken directly from the published TabArena leaderboard rather than re-run in our own environment. 
We omit AutoGluon from the head-to-head comparison as it is a multi-model AutoML stack rather than a single tabular model. All ranks reported below are computed within a $64$-method comparison universe: the $61$ external baselines from the TabArena-Lite leaderboard (the full leaderboard contains $64$ methods including AutoGluon's three AutoML-stack entries, AutoGluon~1.5~(extreme,~4h), AutoGluon~1.4~(extreme,~4h), and AutoGluon~1.4~(best,~4h), which we exclude) plus the three \specmethod{} variants (D), (T), and (T+E). The complete $67$-row leaderboard, including the AutoGluon entries, is reproduced in Appendix~\ref{app:leaderboard}.

\paragraph{Hyperparameter optimization.}
We use a lightweight hyperparameter search consisting of $11$ configurations in total: 1 default configuration and 10 additional random trials. The search space covers the main adapter and optimization choices, including the number of layers, low-rank ratio, BatchNorm, learning rate, initial gate value $\alpha_0$, the learning-rate multiplier for $\alpha$, optimizer choice (AdamW or Muon), preprocessing variant, and learning-rate schedule. The full set of ranges and default values is listed in Appendix~\ref{app:hpo}. In this main sweep, the inner block is fixed to the DCNv2 cross block so that the search budget is concentrated on tuning the default instantiation of the method. The cross-versus-MLP comparison is treated separately as a focused ablation in Appendix~\ref{app:ablations}.

We report three evaluation protocols. \textbf{(D)} uses the default configuration without hyperparameter tuning. \textbf{(T)} selects the single best configuration for each dataset from the 11-trial search. \textbf{(T+E)} further ensembles the top configurations across the 8 validation folds.

Note that our HPO budget is deliberately small. TabArena's published leaderboard tunes each non-Retouche baseline with $200$ random configurations per dataset (on top of a default), so every T or T+E peer we compare against (RealTabPFN-2.5, RealMLP, TabM, TabDPT, the GBDT suite, and others) reflects a $20\times$ larger random-search budget than \specmethod{} receives ($10$ random trials on top of a default). The \specmethod{} (T) and (T+E) numbers reported in Section~\ref{sec:exp:main} are therefore obtained under a substantially constrained search budget. We view them as a lower bound on what the framework can deliver, and expect a budget-matched sweep to widen the reported gains.

\subsection{Main results}
\label{sec:exp:main}

Table~\ref{tab:main} reports the simplified TabArena-Lite leaderboard. \specmethod{} (T+E) reaches Elo $1651$, a $+56$ gain over the unmodified \TabICL{}v2 base ($1595$), and take rank $1$ on the leaderboard. The single-best-config variant \specmethod{} (T) follows at Elo $1632$ (rank $2$, $+37$ vs.\ base). The next-best non-Retouche peers, \TabPFN{}-2.6 (D) and RealTabPFN-2.5 (T+E), sit at Elo $1627$ and $1614$, respectively. \method{}, instantiated on \TabICL{}v2, takes the top two slots of the leaderboard while modifying no pretrained weight and adding only $10^2$--$10^5$ trainable parameters per dataset.

\begin{table}[t]
\caption{Simplified TabArena-Lite leaderboard. 
\specmethod{}'s T and T+E variants use a $20\times$ smaller per-dataset HPO budget than every tuned peer ($10$ random configurations vs.\ TabArena's standard $200$). Time columns are wall-clock per $1$K samples. Bold rows mark \specmethod{} variants. Italic row marks the unmodified TabICLv2.}
\label{tab:main}
\centering
\small
\begin{tabular}{lccccc}
\toprule
\textbf{Method} & \textbf{Elo ($\uparrow$)} & \textbf{Norm. ($\uparrow$)} & \textbf{Avg.\ rank ($\downarrow$)} & \textbf{Train [s/K]} & \textbf{Pred.\ [s/K]} \\
\midrule
\textbf{\specmethod{} (T+E)}            & $\mathbf{1651_{-63,+85}}$ & $\mathbf{0.665}$ & $\mathbf{10.2}$ & $\mathbf{243.17}$ & $\mathbf{22.03}$ \\
\textbf{\specmethod{} (T)}              & $\mathbf{1632_{-61,+74}}$ & $\mathbf{0.659}$ & $\mathbf{11.0}$ & $\mathbf{243.17}$ & $\mathbf{7.33}$ \\
\TabPFN{}-2.6 (D)                       & $1627_{-61,+82}$ & 0.643 & 11.3 & 5.75 & 0.60 \\
RealTabPFN-2.5 (T+E)                    & $1614_{-66,+88}$ & 0.630 & 11.8 & 2059.94 & 9.79 \\
\textbf{\specmethod{} (D)}              & $\mathbf{1608_{-62,+67}}$ & $\mathbf{0.623}$ & $\mathbf{12.1}$ & $\mathbf{20.80}$ & $\mathbf{7.24}$ \\
\TabICL{}v2 (D) & \emph{$1595_{-67,+89}$} & \emph{0.633} & \emph{12.7} & \emph{4.01} & \emph{0.35} \\
RealMLP (T+E)                           & $1500_{-44,+54}$ & 0.458 & 17.7 & 2791.97 & 13.89 \\
LightGBM (T+E)                          & $1400_{-41,+41}$ & 0.275 & 24.0 & 416.56 & 2.24 \\
CatBoost (T+E)                          & $1388_{-45,+60}$ & 0.291 & 24.7 & 1665.53 & 0.56 \\
XGBoost (T+E)                           & $1360_{-46,+43}$ & 0.231 & 26.7 & 700.96 & 1.44 \\
\TabPFN{}v2 (T+E)                       & $1328_{-72,+75}$ & 0.312 & 28.9 & 2942.08 & 17.37 \\
\bottomrule
\end{tabular}
\end{table}

Figure~\ref{fig:tuning-impact} plots, for each method that ships both a default and a tuned variant, the Elo gain from per-dataset hyperparameter tuning. Unlike baselines whose defaults sit far below their tuned ceiling (e.g.\ RealMLP, LightGBM, and XGBoost each gain $>$$170$ Elo from D to T+E), \specmethod{} starts from a strong default at Elo $1608$ that already exceeds the unmodified \TabICL{}v2 base, and tuning further lifts it by $+43$ Elo to $1651$ at rank $1$. This demonstrates that \specmethod{} delivers a useful lift either out of the box or with a small per-task search.

\begin{figure}[t]
\centering
\begin{subfigure}{0.49\linewidth}
  \centering
  \includegraphics[width=\linewidth]{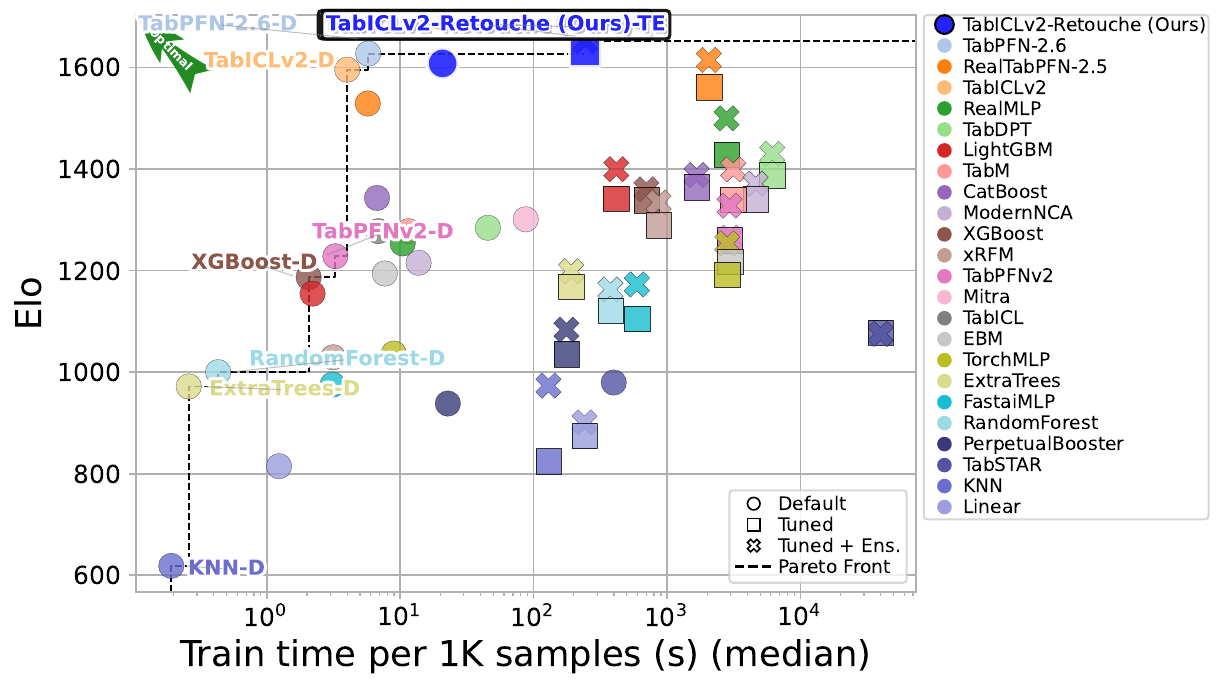}
  \caption{Elo vs.\ training time per $1$K samples.}
  \label{fig:pareto_train_elo}
\end{subfigure}\hfill
\begin{subfigure}{0.49\linewidth}
  \centering
  \includegraphics[width=\linewidth]{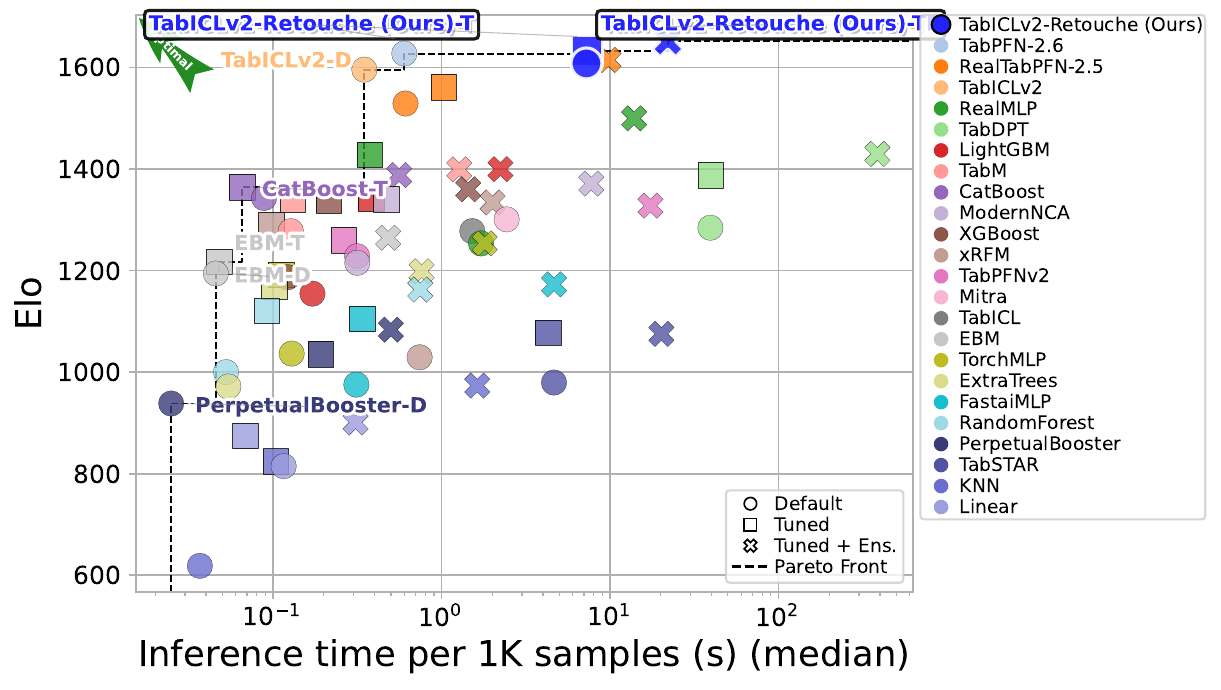}
  \caption{Elo vs.\ inference time per $1$K samples.}
  \label{fig:pareto_infer_elo}
\end{subfigure}
\caption{Quality vs. compute trade-off on TabArena-Lite:  Elo vs.\ training time (a) and Elo vs.\ inference time (b). Upper-left is better; markers denote leaderboard methods. On the training-time axis (a), \specmethod{} (T+E) is the highest-Elo point on the Pareto frontier and dominates every other tuned-and-ensembled peer. On the inference-time axis (b), \specmethod{} (T+E) is the highest-Elo point on the frontier; \specmethod{} (T) lies on the same frontier as a lower-overhead variant.}
\label{fig:pareto_elo}
\end{figure}

Figure~\ref{fig:pareto_elo} plots Elo against per-thousand-sample training time (left) and inference time (right) for every method on the TabArena-Lite leaderboard. Appendix~\ref{app:headline-extras} shows the same data with improvability on the y-axis (Figure~\ref{fig:pareto}) and the corresponding critical-difference diagram (Figure~\ref{fig:critical-diagram}). On the training-time axis, \specmethod{} (T+E) sits on the Pareto frontier as the highest-Elo operating point at $243$s per $1$K samples. It dominates every other (T+E) peer on both quality and training time (e.g. RealTabPFN-2.5 (T+E) at $2{,}060$s, RealMLP (T+E) at $2{,}792$s). \TabPFN{}-2.6 (D) and the unmodified \TabICL{}v2 (D) cover the fast end of the same frontier. On the inference-time axis, \specmethod{} (T+E) at $22$s/K is the highest-Elo point on the frontier, with \specmethod{} (T) at $7.3$s/K and the unmodified \TabICL{}v2 / \TabPFN{}-2.6 defaults covering the lower-Elo, faster-inference end. Here \specmethod{} (T+E) trades inference time against quality rather than dominating the (T+E) peers outright. The headline finding is therefore that a small input-space adapter on a frozen TFM tops the TabArena-Lite leaderboard, dominates the training-time Pareto frontier among tuned-and-ensembled methods, and remains on the inference-time frontier.

\specmethod{}'s wall-clock numbers were measured on a mix of NVIDIA A10 and A100 GPUs, whereas TabArena's published times for GPU-based baselines (\TabICL{}v2, \TabPFN{}-2.6, RealTabPFN-2.5, TabM, TabDPT) were obtained on faster H100s. The training-time dominance and the inference-time Pareto-frontier position therefore hold \emph{despite} a hardware handicap. Under matched hardware the time numbers would only shift further toward the fast end of the plots.

We extended the evaluation to the TALENT benchmark~\citep{liu_talent_2024}. The result is shown in Appendix~\ref{app:talent}

\section{Limitations and Future Work}
\label{sec:limits}

Several choices in this paper trade absolute performance for tractability under a constrained compute budget. First, our per-dataset random search uses only $10$ configurations on top of one default, $20\times$ smaller than TabArena's published $200$-trial sweeps. Second, we share a single HPO search space across classification and regression rather than tuning per problem type, even though HPO that helps one can hurt the other~\citep{holzmuller_better_2025}. Problem-type-conditioned defaults or meta-learned configurations are a natural follow-up. Third, the reported wall-clock numbers were measured on NVIDIA A10 and A100 GPUs, while TabArena's published baseline times use faster H100s. Each of these factors systematically biases the reported \specmethod{} results downward.
We therefore view the reported leaderboard position and Pareto-frontier dominance as a lower bound on what \specmethod{} can deliver under matched conditions, rather than as a reflection of the framework’s full potential.

We benchmark only on \TabICL{}v2. \TabPFN{} models (\TabPFN{}-2.5/2.6) are not yet evaluated because of the license. \method{} is backbone-agnostic by design, and porting to other openly licensed TFMs is straightforward in principle but not yet evaluated.  

In this paper we deliberately used a simple implementation to focus on the idea of \method{}: gated residual adapter with identity fallback.  
Several orthogonal ideas from recent tabular work fit naturally on top of this skeleton: richer per-feature numeric encodings such as PLR~\citep{gorishniy_embeddings_2023} together with the robust preprocessing pipeline and better optimization tricks in RealMLP~\citep{holzmuller_better_2025} to give the residual branch a stronger input representation; TabM-style BatchEnsemble~\citep{gorishniy_tabm_2025} to turn a single adapter into a low-variance committee at low parameter cost; and BETA-style bagging~\citep{liu_tabpfn_2025} of the trained adapter at inference to further damp prediction variance. We expect each to compose cleanly with the framework, lifting the reported numbers above the conservative single-adapter baseline used here.

\section{Conclusion}
\label{sec:conclusion}

\method{} frames the adaptation of a tabular foundation model as a small, near-identity correction in \emph{input space}: a learnable gate combined with a lightweight inner block ($10^2$--$10^5$ trainable parameters per dataset, scaling with input dimensionality), trained end-to-end through the frozen backbone, and gated at deployment by a post-training identity check on held-out validation. No pretrained weight is modified and no TFM internal-architecture choices are required, leaving the backbone a swappable knob rather than a structural commitment. 

Empirically, \specmethod{}---the framework instantiated on \TabICL{}v2---is the rank-$1$ method on TabArena-Lite at Elo $1651$ (T+E, $+56$ over the frozen base), dominates the training-time Pareto frontier among tuned-and-ensembled methods, and remains on the inference-time Pareto frontier, showing that input-space adaptation can deliver state-of-the-art quality without sacrificing efficiency. These gains are obtained with only $10$ random HPO trials per dataset---$20\times$ smaller than the $200$-trial budget behind the published TabArena baselines.

\method{} is intended as a late-stage refinement rather than a replacement for the zero-shot TFM workflow. Modern TFMs already deliver competitive predictions with no per-dataset training, which makes them well-suited to fast iteration on the upstream decisions---data sources, feature schema, data representation, evaluation, monitoring---where most early-stage data-science effort belongs and where the frozen TFM is hard to beat as a default predictor. Once the upstream pipeline has stabilized, marginal predictive gains start to translate into real-world value, either as raw accuracy or as personalization toward a target distribution, and \method{} provides a small, low-friction mechanism for capturing them. \method{} leaves the frozen backbone untouched so the existing zero-shot pipeline remains the fallback. The per-dataset adapter ($10^2$--$10^5$ trainable parameters) trains on a single GPU at adapter-only cost rather than backbone fine-tuning cost. The identity guard (Section~\ref{sec:method:guard}) routes back to the unmodified TFM on any dataset where adaptation does not improve held-out performance, making the addition strictly opt-in per dataset. And because the adapter is architecture-agnostic with respect to the backbone, the choice of TFM remains a swappable knob. This facilitate the adoption of \method{}, because it can incorporate future stronger TFMs naturally. Together these properties position \method{} as a low-risk, late-stage performance lift after the upstream data and pipeline decisions have stabilized.


\begin{ack}
We thank David Holzmüller for valuable discussions on deep learning and optimization techniques for tabular models, and for his advice on the choice of baselines against which to compare \method{}.
\end{ack}

\bibliographystyle{plainnat}
\bibliography{references}

@misc{loshchilov_decoupled_2019,
	title = {Decoupled {Weight} {Decay} {Regularization}},
	url = {http://arxiv.org/abs/1711.05101},
	doi = {10.48550/arXiv.1711.05101},
	urldate = {2026-05-06},
	publisher = {arXiv},
	author = {Loshchilov, Ilya and Hutter, Frank},
	month = jan,
	year = {2019},
	note = {arXiv:1711.05101 [cs]},
}

@misc{bachlechner_rezero_2020,
	title = {{ReZero} is {All} {You} {Need}: {Fast} {Convergence} at {Large} {Depth}},
	shorttitle = {{ReZero} is {All} {You} {Need}},
	url = {http://arxiv.org/abs/2003.04887},
	doi = {10.48550/arXiv.2003.04887},
	urldate = {2026-05-03},
	publisher = {arXiv},
	author = {Bachlechner, Thomas and Majumder, Bodhisattwa Prasad and Mao, Huanru Henry and Cottrell, Garrison W. and McAuley, Julian},
	month = jun,
	year = {2020},
	note = {arXiv:2003.04887 [cs]},
}

@misc{touvron_going_2021,
	title = {Going deeper with {Image} {Transformers}},
	url = {http://arxiv.org/abs/2103.17239},
	doi = {10.48550/arXiv.2103.17239},
	urldate = {2026-05-03},
	publisher = {arXiv},
	author = {Touvron, Hugo and Cord, Matthieu and Sablayrolles, Alexandre and Synnaeve, Gabriel and Jégou, Hervé},
	month = apr,
	year = {2021},
	note = {arXiv:2103.17239 [cs]},
}

@misc{gorishniy_embeddings_2023,
	title = {On {Embeddings} for {Numerical} {Features} in {Tabular} {Deep} {Learning}},
	url = {http://arxiv.org/abs/2203.05556},
	doi = {10.48550/arXiv.2203.05556},
	urldate = {2026-04-30},
	publisher = {arXiv},
	author = {Gorishniy, Yury and Rubachev, Ivan and Babenko, Artem},
	month = oct,
	year = {2023},
	note = {arXiv:2203.05556 [cs]},
}

@misc{grinsztajn_tabpfn-25_2025,
	title = {{TabPFN}-2.5},
	author = {Grinsztajn, Léo and Flöge, Klemens and Key, Oscar and Birkel, Felix and Roof, Brendan and Jund, Phil and Jäger, Benjamin and Hayler, Adrian and Safaric, Dominik and Simone Alessi, Felix Jablonski and Manium, Mihir and Yu, Rosen and Garg, Anurag and Robertson, Jake and Hoo, Shi Bin (Liam) and Moroshan, Vladyslav and Bühler, Magnus and Purucker, Lennart and Cornu, Clara and Wehrhahn, Lilly Charlotte and Bonetto, Alessandro and Gambhir, Sauraj and Hollmann, Noah and Hutter, Frank},
	year = {2025},
}

@article{dastile_statistical_2020,
	title = {Statistical and machine learning models in credit scoring: {A} systematic literature survey},
	volume = {91},
	issn = {1568-4946},
	shorttitle = {Statistical and machine learning models in credit scoring},
	url = {https://www.sciencedirect.com/science/article/pii/S1568494620302039},
	doi = {10.1016/j.asoc.2020.106263},
	urldate = {2026-04-30},
	journal = {Applied Soft Computing},
	author = {Dastile, Xolani and Celik, Turgay and Potsane, Moshe},
	month = jun,
	year = {2020},
	pages = {106263},
}

@inproceedings{chen_xgboost_2016,
	address = {San Francisco California USA},
	title = {{XGBoost}: {A} {Scalable} {Tree} {Boosting} {System}},
	isbn = {978-1-4503-4232-2},
	shorttitle = {{XGBoost}},
	url = {https://dl.acm.org/doi/10.1145/2939672.2939785},
	doi = {10.1145/2939672.2939785},
	language = {en},
	urldate = {2026-04-30},
	booktitle = {Proceedings of the 22nd {ACM} {SIGKDD} {International} {Conference} on {Knowledge} {Discovery} and {Data} {Mining}},
	publisher = {ACM},
	author = {Chen, Tianqi and Guestrin, Carlos},
	month = aug,
	year = {2016},
	pages = {785--794},
}

@inproceedings{ke_lightgbm_2017,
	title = {{LightGBM}: {A} {Highly} {Efficient} {Gradient} {Boosting} {Decision} {Tree}},
	volume = {30},
	shorttitle = {{LightGBM}},
	url = {https://proceedings.neurips.cc/paper_files/paper/2017/hash/6449f44a102fde848669bdd9eb6b76fa-Abstract.html},
	urldate = {2026-04-30},
	booktitle = {Advances in {Neural} {Information} {Processing} {Systems}},
	publisher = {Curran Associates, Inc.},
	author = {Ke, Guolin and Meng, Qi and Finley, Thomas and Wang, Taifeng and Chen, Wei and Ma, Weidong and Ye, Qiwei and Liu, Tie-Yan},
	year = {2017},
}

@article{shwartz-ziv_tabular_2022,
	title = {Tabular data: {Deep} learning is not all you need},
	volume = {81},
	issn = {15662535},
	shorttitle = {Tabular data},
	url = {https://linkinghub.elsevier.com/retrieve/pii/S1566253521002360},
	doi = {10.1016/j.inffus.2021.11.011},
	language = {en},
	urldate = {2026-04-30},
	journal = {Information Fusion},
	author = {Shwartz-Ziv, Ravid and Armon, Amitai},
	month = may,
	year = {2022},
	pages = {84--90},
}

@article{borisov_deep_2024,
	title = {Deep {Neural} {Networks} and {Tabular} {Data}: {A} {Survey}},
	volume = {35},
	copyright = {https://creativecommons.org/licenses/by/4.0/legalcode},
	issn = {2162-237X, 2162-2388},
	shorttitle = {Deep {Neural} {Networks} and {Tabular} {Data}},
	url = {https://ieeexplore.ieee.org/document/9998482/},
	doi = {10.1109/TNNLS.2022.3229161},
	language = {en},
	number = {6},
	urldate = {2026-04-30},
	journal = {IEEE Transactions on Neural Networks and Learning Systems},
	author = {Borisov, Vadim and Leemann, Tobias and Seßler, Kathrin and Haug, Johannes and Pawelczyk, Martin and Kasneci, Gjergji},
	month = jun,
	year = {2024},
	pages = {7499--7519},
}

@article{fatima_survey_2017,
	title = {Survey of {Machine} {Learning} {Algorithms} for {Disease} {Diagnostic}},
	volume = {9},
	copyright = {https://creativecommons.org/licenses/by/4.0/},
	url = {https://www.scirp.org/journal/paperinformation?paperid=73781},
	doi = {10.4236/jilsa.2017.91001},
	language = {en},
	number = {1},
	urldate = {2026-04-30},
	journal = {Journal of Intelligent Learning Systems and Applications},
	publisher = {Scientific Research Publishing},
	author = {Fatima, Meherwar and Pasha, Maruf},
	month = jan,
	year = {2017},
	pages = {1--16},
}

@misc{breugel_why_2024,
	title = {Why {Tabular} {Foundation} {Models} {Should} {Be} a {Research} {Priority}},
	url = {http://arxiv.org/abs/2405.01147},
	doi = {10.48550/arXiv.2405.01147},
	urldate = {2026-04-30},
	publisher = {arXiv},
	author = {Breugel, Boris van and Schaar, Mihaela van der},
	month = jun,
	year = {2024},
	note = {arXiv:2405.01147 [cs]},
}

@misc{rubachev_finetuning_2025,
	title = {On {Finetuning} {Tabular} {Foundation} {Models}},
	copyright = {Creative Commons Attribution Non Commercial Share Alike 4.0 International},
	url = {https://arxiv.org/abs/2506.08982},
	doi = {10.48550/ARXIV.2506.08982},
	language = {en},
	urldate = {2026-04-29},
	publisher = {arXiv},
	author = {Rubachev, Ivan and Kotelnikov, Akim and Kartashev, Nikolay and Babenko, Artem},
	year = {2025},
	note = {Version Number: 2},
}

@misc{bouadi_orion-msp_2025,
	title = {Orion-{MSP}: {Multi}-{Scale} {Sparse} {Attention} for {Tabular} {In}-{Context} {Learning}},
	shorttitle = {Orion-{MSP}},
	url = {http://arxiv.org/abs/2511.02818},
	doi = {10.48550/arXiv.2511.02818},
	urldate = {2026-04-29},
	publisher = {arXiv},
	author = {Bouadi, Mohamed and Seth, Pratinav and Tanna, Aditya and Sankarapu, Vinay Kumar},
	month = nov,
	year = {2025},
	note = {arXiv:2511.02818 [cs]},
}

@misc{zhang_limix_2025,
	title = {{LimiX}: {Unleashing} {Structured}-{Data} {Modeling} {Capability} for {Generalist} {Intelligence}},
	shorttitle = {{LimiX}},
	url = {http://arxiv.org/abs/2509.03505},
	doi = {10.48550/arXiv.2509.03505},
	urldate = {2026-04-29},
	publisher = {arXiv},
	author = {Zhang, Xingxuan and Ren, Gang and Yu, Han and Yuan, Hao and Wang, Hui and Li, Jiansheng and Wu, Jiayun and Mo, Lang and Mao, Li and Hao, Mingchao and Dai, Ningbo and Xu, Renzhe and Li, Shuyang and Zhang, Tianyang and He, Yue and Wang, Yuanrui and Zhang, Yunjia and Xu, Zijing and Li, Dongzhe and Gao, Fang and Zou, Hao and Liu, Jiandong and Liu, Jiashuo and Xu, Jiawei and Cheng, Kaijie and Li, Kehan and Zhou, Linjun and Li, Qing and Fan, Shaohua and Lin, Xiaoyu and Han, Xinyan and Li, Xuanyue and Lu, Yan and Xue, Yuan and Jiang, Yuanyuan and Wang, Zimu and Wang, Zhenlei and Cui, Peng},
	month = nov,
	year = {2025},
	note = {arXiv:2509.03505 [cs]},
}

@misc{nori_interpretml_2019,
	title = {{InterpretML}: {A} {Unified} {Framework} for {Machine} {Learning} {Interpretability}},
	shorttitle = {{InterpretML}},
	url = {http://arxiv.org/abs/1909.09223},
	doi = {10.48550/arXiv.1909.09223},
	urldate = {2026-04-28},
	publisher = {arXiv},
	author = {Nori, Harsha and Jenkins, Samuel and Koch, Paul and Caruana, Rich},
	month = sep,
	year = {2019},
	note = {arXiv:1909.09223 [cs]},
}

@misc{ye_revisiting_2025,
	title = {Revisiting {Nearest} {Neighbor} for {Tabular} {Data}: {A} {Deep} {Tabular} {Baseline} {Two} {Decades} {Later}},
	shorttitle = {Revisiting {Nearest} {Neighbor} for {Tabular} {Data}},
	url = {http://arxiv.org/abs/2407.03257},
	doi = {10.48550/arXiv.2407.03257},
	urldate = {2026-04-28},
	publisher = {arXiv},
	author = {Ye, Han-Jia and Yin, Huai-Hong and Zhan, De-Chuan and Chao, Wei-Lun},
	month = mar,
	year = {2025},
	note = {arXiv:2407.03257 [cs]},
}

@misc{liu_talent_2024,
	title = {{TALENT}: {A} {Tabular} {Analytics} and {Learning} {Toolbox}},
	shorttitle = {{TALENT}},
	url = {http://arxiv.org/abs/2407.04057},
	doi = {10.48550/arXiv.2407.04057},
	urldate = {2026-04-27},
	publisher = {arXiv},
	author = {Liu, Si-Yang and Cai, Hao-Run and Zhou, Qi-Le and Ye, Han-Jia},
	month = jul,
	year = {2024},
	note = {arXiv:2407.04057 [cs]},
}

@misc{ma_tabdpt_2026,
	title = {{TabDPT}: {Scaling} {Tabular} {Foundation} {Models} on {Real} {Data}},
	shorttitle = {{TabDPT}},
	url = {http://arxiv.org/abs/2410.18164},
	doi = {10.48550/arXiv.2410.18164},
	urldate = {2026-04-27},
	publisher = {arXiv},
	author = {Ma, Junwei and Thomas, Valentin and Hosseinzadeh, Rasa and Labach, Alex and Kamkari, Hamidreza and Cresswell, Jesse C. and Golestan, Keyvan and Yu, Guangwei and Caterini, Anthony L. and Volkovs, Maksims},
	month = jan,
	year = {2026},
	note = {arXiv:2410.18164 [cs]},
}

@misc{zhang_mitra_2025,
	title = {Mitra: {Mixed} {Synthetic} {Priors} for {Enhancing} {Tabular} {Foundation} {Models}},
	shorttitle = {Mitra},
	url = {http://arxiv.org/abs/2510.21204},
	doi = {10.48550/arXiv.2510.21204},
	urldate = {2026-04-27},
	publisher = {arXiv},
	author = {Zhang, Xiyuan and Maddix, Danielle C. and Yin, Junming and Erickson, Nick and Ansari, Abdul Fatir and Han, Boran and Zhang, Shuai and Akoglu, Leman and Faloutsos, Christos and Mahoney, Michael W. and Hu, Cuixiong and Rangwala, Huzefa and Karypis, George and Wang, Bernie},
	month = oct,
	year = {2025},
	note = {arXiv:2510.21204 [cs]},
}

@misc{kim_carte_2024,
	title = {{CARTE}: {Pretraining} and {Transfer} for {Tabular} {Learning}},
	shorttitle = {{CARTE}},
	url = {http://arxiv.org/abs/2402.16785},
	doi = {10.48550/arXiv.2402.16785},
	urldate = {2026-04-27},
	publisher = {arXiv},
	author = {Kim, Myung Jun and Grinsztajn, Léo and Varoquaux, Gaël},
	month = may,
	year = {2024},
	note = {arXiv:2402.16785 [cs]},
}

@misc{tanna_exploring_2026,
	title = {Exploring {Fine}-{Tuning} for {Tabular} {Foundation} {Models}},
	url = {http://arxiv.org/abs/2601.09654},
	doi = {10.48550/arXiv.2601.09654},
	urldate = {2026-04-27},
	publisher = {arXiv},
	author = {Tanna, Aditya and Seth, Pratinav and Bouadi, Mohamed and Sankarapu, Vinay Kumar},
	month = jan,
	year = {2026},
	note = {arXiv:2601.09654 [cs]},
}

@misc{tanna_tabtune_2025,
	title = {{TabTune}: {A} {Unified} {Library} for {Inference} and {Fine}-{Tuning} {Tabular} {Foundation} {Models}},
	shorttitle = {{TabTune}},
	url = {http://arxiv.org/abs/2511.02802},
	doi = {10.48550/arXiv.2511.02802},
	urldate = {2026-04-27},
	publisher = {arXiv},
	author = {Tanna, Aditya and Seth, Pratinav and Bouadi, Mohamed and Avaiya, Utsav and Sankarapu, Vinay Kumar},
	month = dec,
	year = {2025},
	note = {arXiv:2511.02802 [cs]},
}

@misc{jordan_muon_2024,
	title = {Muon: {An} optimizer for hidden layers in neural networks},
	url = {https://kellerjordan.github.io/posts/muon/},
	author = {Jordan, Keller and Jin, Yuchen and Boza, Vlado and Jiacheng, You and Cesista, Franz and Newhouse, Laker and Bernstein, Jeremy},
	year = {2024},
}

@misc{spinaci_contexttab_2025,
	title = {{ConTextTab}: {A} {Semantics}-{Aware} {Tabular} {In}-{Context} {Learner}},
	shorttitle = {{ConTextTab}},
	url = {http://arxiv.org/abs/2506.10707},
	doi = {10.48550/arXiv.2506.10707},
	urldate = {2026-04-27},
	publisher = {arXiv},
	author = {Spinaci, Marco and Polewczyk, Marek and Schambach, Maximilian and Thelin, Sam},
	month = nov,
	year = {2025},
	note = {arXiv:2506.10707 [cs]},
}

@misc{arazi_tabstar_2025,
	title = {{TabSTAR}: {A} {Foundation} {Tabular} {Model} {With} {Semantically} {Target}-{Aware} {Representations}},
	shorttitle = {{TabSTAR}},
	url = {http://arxiv.org/abs/2505.18125},
	doi = {10.48550/arXiv.2505.18125},
	urldate = {2026-05-08},
	publisher = {arXiv},
	author = {Arazi, Alan and Shapira, Eilam and Reichart, Roi},
	month = may,
	year = {2025},
	note = {arXiv:2505.18125 [cs]},
}

@misc{erickson_tabarena_2025,
	title = {{TabArena}: {A} {Living} {Benchmark} for {Machine} {Learning} on {Tabular} {Data}},
	shorttitle = {{TabArena}},
	url = {http://arxiv.org/abs/2506.16791},
	doi = {10.48550/arXiv.2506.16791},
	urldate = {2026-04-27},
	publisher = {arXiv},
	author = {Erickson, Nick and Purucker, Lennart and Tschalzev, Andrej and Holzmüller, David and Desai, Prateek Mutalik and Salinas, David and Hutter, Frank},
	month = nov,
	year = {2025},
	note = {arXiv:2506.16791 [cs]},
}

@misc{gorishniy_tabm_2025,
	title = {{TabM}: {Advancing} {Tabular} {Deep} {Learning} with {Parameter}-{Efficient} {Ensembling}},
	shorttitle = {{TabM}},
	url = {http://arxiv.org/abs/2410.24210},
	doi = {10.48550/arXiv.2410.24210},
	urldate = {2026-04-27},
	publisher = {arXiv},
	author = {Gorishniy, Yury and Kotelnikov, Akim and Babenko, Artem},
	month = feb,
	year = {2025},
	note = {arXiv:2410.24210 [cs]},
}

@misc{prokhorenkova_catboost_2019,
	title = {{CatBoost}: unbiased boosting with categorical features},
	shorttitle = {{CatBoost}},
	url = {http://arxiv.org/abs/1706.09516},
	doi = {10.48550/arXiv.1706.09516},
	urldate = {2026-04-27},
	publisher = {arXiv},
	author = {Prokhorenkova, Liudmila and Gusev, Gleb and Vorobev, Aleksandr and Dorogush, Anna Veronika and Gulin, Andrey},
	month = jan,
	year = {2019},
	note = {arXiv:1706.09516 [cs]},
}

@misc{holzmuller_better_2025,
	title = {Better by {Default}: {Strong} {Pre}-{Tuned} {MLPs} and {Boosted} {Trees} on {Tabular} {Data}},
	shorttitle = {Better by {Default}},
	url = {http://arxiv.org/abs/2407.04491},
	doi = {10.48550/arXiv.2407.04491},
	urldate = {2026-04-27},
	publisher = {arXiv},
	author = {Holzmüller, David and Grinsztajn, Léo and Steinwart, Ingo},
	month = jan,
	year = {2025},
	note = {arXiv:2407.04491 [cs]},
}

@misc{hu_lora_2021,
	title = {{LoRA}: {Low}-{Rank} {Adaptation} of {Large} {Language} {Models}},
	shorttitle = {{LoRA}},
	url = {http://arxiv.org/abs/2106.09685},
	doi = {10.48550/arXiv.2106.09685},
	urldate = {2026-04-27},
	publisher = {arXiv},
	author = {Hu, Edward J. and Shen, Yelong and Wallis, Phillip and Allen-Zhu, Zeyuan and Li, Yuanzhi and Wang, Shean and Wang, Lu and Chen, Weizhu},
	month = oct,
	year = {2021},
	note = {arXiv:2106.09685 [cs]},
}

@misc{liu_tabpfn_2025,
	title = {{TabPFN} {Unleashed}: {A} {Scalable} and {Effective} {Solution} to {Tabular} {Classification} {Problems}},
	shorttitle = {{TabPFN} {Unleashed}},
	url = {http://arxiv.org/abs/2502.02527},
	doi = {10.48550/arXiv.2502.02527},
	urldate = {2026-04-27},
	publisher = {arXiv},
	author = {Liu, Si-Yang and Ye, Han-Jia},
	month = feb,
	year = {2025},
	note = {arXiv:2502.02527 [cs]},
}

@misc{grinsztajn_tabpfn-25_2026,
	title = {{TabPFN}-2.5: {Advancing} the {State} of the {Art} in {Tabular} {Foundation} {Models}},
	shorttitle = {{TabPFN}-2.5},
	url = {http://arxiv.org/abs/2511.08667},
	doi = {10.48550/arXiv.2511.08667},
	urldate = {2026-04-27},
	publisher = {arXiv},
	author = {Grinsztajn, Léo and Flöge, Klemens and Key, Oscar and Birkel, Felix and Jund, Philipp and Roof, Brendan and Jäger, Benjamin and Safaric, Dominik and Alessi, Simone and Hayler, Adrian and Manium, Mihir and Yu, Rosen and Jablonski, Felix and Hoo, Shi Bin and Garg, Anurag and Robertson, Jake and Bühler, Magnus and Moroshan, Vladyslav and Purucker, Lennart and Cornu, Clara and Wehrhahn, Lilly Charlotte and Bonetto, Alessandro and Schölkopf, Bernhard and Gambhir, Sauraj and Hollmann, Noah and Hutter, Frank},
	month = feb,
	year = {2026},
	note = {arXiv:2511.08667 [cs]},
}

@inproceedings{wang_dcn_2021,
	title = {{DCN} {V2}: {Improved} {Deep} \& {Cross} {Network} and {Practical} {Lessons} for {Web}-scale {Learning} to {Rank} {Systems}},
	shorttitle = {{DCN} {V2}},
	url = {http://arxiv.org/abs/2008.13535},
	doi = {10.1145/3442381.3450078},
	urldate = {2026-04-27},
	booktitle = {Proceedings of the {Web} {Conference} 2021},
	author = {Wang, Ruoxi and Shivanna, Rakesh and Cheng, Derek Z. and Jain, Sagar and Lin, Dong and Hong, Lichan and Chi, Ed H.},
	month = apr,
	year = {2021},
	note = {arXiv:2008.13535 [cs]},
	pages = {1785--1797},
}

@article{hollmann_accurate_2025,
	title = {Accurate predictions on small data with a tabular foundation model},
	volume = {637},
	copyright = {2025 The Author(s)},
	issn = {1476-4687},
	url = {https://www.nature.com/articles/s41586-024-08328-6},
	doi = {10.1038/s41586-024-08328-6},
	language = {en},
	number = {8045},
	urldate = {2026-04-27},
	journal = {Nature},
	publisher = {Nature Publishing Group},
	author = {Hollmann, Noah and Müller, Samuel and Purucker, Lennart and Krishnakumar, Arjun and Körfer, Max and Hoo, Shi Bin and Schirrmeister, Robin Tibor and Hutter, Frank},
	month = jan,
	year = {2025},
	pages = {319--326},
}

@misc{hollmann_tabpfn_2023,
	title = {{TabPFN}: {A} {Transformer} {That} {Solves} {Small} {Tabular} {Classification} {Problems} in a {Second}},
	shorttitle = {{TabPFN}},
	url = {http://arxiv.org/abs/2207.01848},
	doi = {10.48550/arXiv.2207.01848},
	urldate = {2026-04-27},
	publisher = {arXiv},
	author = {Hollmann, Noah and Müller, Samuel and Eggensperger, Katharina and Hutter, Frank},
	month = sep,
	year = {2023},
	note = {arXiv:2207.01848 [cs]},
}

@misc{qu_tabicl_2025,
	title = {{TabICL}: {A} {Tabular} {Foundation} {Model} for {In}-{Context} {Learning} on {Large} {Data}},
	shorttitle = {{TabICL}},
	url = {http://arxiv.org/abs/2502.05564},
	doi = {10.48550/arXiv.2502.05564},
	urldate = {2026-04-27},
	publisher = {arXiv},
	author = {Qu, Jingang and Holzmüller, David and Varoquaux, Gaël and Morvan, Marine Le},
	month = may,
	year = {2025},
	note = {arXiv:2502.05564 [cs]},
}

@misc{qu_tabiclv2_2026,
	title = {{TabICLv2}: {A} better, faster, scalable, and open tabular foundation model},
	shorttitle = {{TabICLv2}},
	url = {http://arxiv.org/abs/2602.11139},
	doi = {10.48550/arXiv.2602.11139},
	urldate = {2026-04-27},
	publisher = {arXiv},
	author = {Qu, Jingang and Holzmüller, David and Varoquaux, Gaël and Morvan, Marine Le},
	month = feb,
	year = {2026},
	note = {arXiv:2602.11139 [cs]},
}

@article{altman_financial_1968,
	title = {Financial Ratios, Discriminant Analysis and the Prediction of Corporate Bankruptcy},
	author = {Altman, Edward I.},
	journal = {The Journal of Finance},
	volume = {23},
	number = {4},
	pages = {589--609},
	year = {1968},
	publisher = {Wiley/American Finance Association},
}

@article{shumway_forecasting_2001,
	title = {Forecasting Bankruptcy More Accurately: A Simple Hazard Model},
	author = {Shumway, Tyler},
	journal = {The Journal of Business},
	volume = {74},
	number = {1},
	pages = {101--124},
	year = {2001},
	publisher = {The University of Chicago Press},
}

@article{chava_bankruptcy_2004,
	title = {Bankruptcy Prediction with Industry Effects},
	author = {Chava, Sudheer and Jarrow, Robert A.},
	journal = {Review of Finance},
	volume = {8},
	number = {4},
	pages = {537--569},
	year = {2004},
	publisher = {Oxford University Press},
}

@article{ohlson_financial_1980,
	title = {Financial Ratios and the Probabilistic Prediction of Bankruptcy},
	author = {Ohlson, James A.},
	journal = {Journal of Accounting Research},
	volume = {18},
	number = {1},
	pages = {109--131},
	year = {1980},
	publisher = {Wiley},
}

@article{laitinen_bankruptcy_2000,
	title = {Bankruptcy Prediction: Application of the {T}aylor's Expansion in Logistic Regression},
	author = {Laitinen, Erkki K. and Laitinen, Teija},
	journal = {International Review of Financial Analysis},
	volume = {9},
	number = {4},
	pages = {327--349},
	year = {2000},
	publisher = {Elsevier},
}
\medskip



\newpage
\appendix







\section{Equivalent residual form of the gated adapter}
\label{app:residual-form}

Equation~\ref{eq:adapter} writes the adapter as a convex blend of the raw input $x$ and an inner-block output $\delta(x)$ under a learnable gate $\alpha$. We show here that this is algebraically identical to a pure additive residual on $x$ whose magnitude is controlled by $\alpha$, and that the resulting residual has a clean closed form for both inner blocks of Section~\ref{sec:method:blocks}.

Distributing the gate in Equation~\ref{eq:adapter} gives
\begin{align}
\adapter(x) \;=\; (1-\alpha) \odot x + \alpha \odot \delta(x)
            \;=\; x \;-\; \alpha \odot x \;+\; \alpha \odot \delta(x)
            \;=\; x \;+\; \alpha \odot \big(\delta(x) - x\big).
\label{eq:adapter-residual-form}
\end{align}
Defining the inner-block residual
\begin{equation}
r(x) \;:=\; \delta(x) - x,
\label{eq:inner-residual}
\end{equation}
the adapter is therefore a pure additive residual on $x$ with effective gated residual $\alpha \odot r(x)$:
\begin{equation}
\adapter(x) \;=\; x \;+\; \alpha \odot r(x).
\label{eq:adapter-residual-final}
\end{equation}

Because both inner blocks defined in Section~\ref{sec:method:blocks} are themselves residual stacks initialized at $x_0 = x$, the inner-block residual $r(x) = \delta(x) - x$ telescopes into a sum of per-layer increments and is available in closed form.

For the cross block (Equation~\ref{eq:cross-block}), $x_{l+1} = x_l + x_0 \odot (W_l x_l + b_l)$, so
\begin{equation}
r_{\text{cross}}(x) \;=\; \delta_{\text{cross}}(x) - x \;=\; \sum_{l=0}^{L-1} x_0 \odot (W_l x_l + b_l),
\label{eq:cross-residual}
\end{equation}
which makes explicit that the cross-block contribution is a layer-wise sum of input-anchored multiplicative corrections. 

For the residual-MLP block (Equation~\ref{eq:mlp-block}), $x_{l+1} = x_l + W_l^{2} \sigma(W_l^{1} x_l + b_l^{1}) + b_l^{2}$, so
\begin{equation}
r_{\text{MLP}}(x) \;=\; \delta_{\text{MLP}}(x) - x \;=\; \sum_{l=0}^{L-1} \Big( W_l^{2} \sigma(W_l^{1} x_l + b_l^{1}) + b_l^{2} \Big),
\label{eq:mlp-residual}
\end{equation}
i.e.\ a layer-wise sum of bottlenecked nonlinear corrections. In both cases, substituting Equation~\ref{eq:cross-residual} or Equation~\ref{eq:mlp-residual} into Equation~\ref{eq:adapter-residual-final} yields an explicit per-channel residual whose overall magnitude is controlled by the learned gate $\alpha$.


\section{Additional headline-run results}
\label{app:headline-extras}

This appendix collects two complementary views of the headline cross-block run reported in Section~\ref{sec:exp:main}: a quality-vs-compute Pareto plot under the alternative \emph{improvability} y-axis, and a Demšar-style critical-difference diagram that ranks the headline run against every non-AutoGluon TabArena-Lite leaderboard baseline.

\begin{figure}[t]
\centering
\begin{subfigure}{0.49\linewidth}
  \centering
  \includegraphics[width=\linewidth]{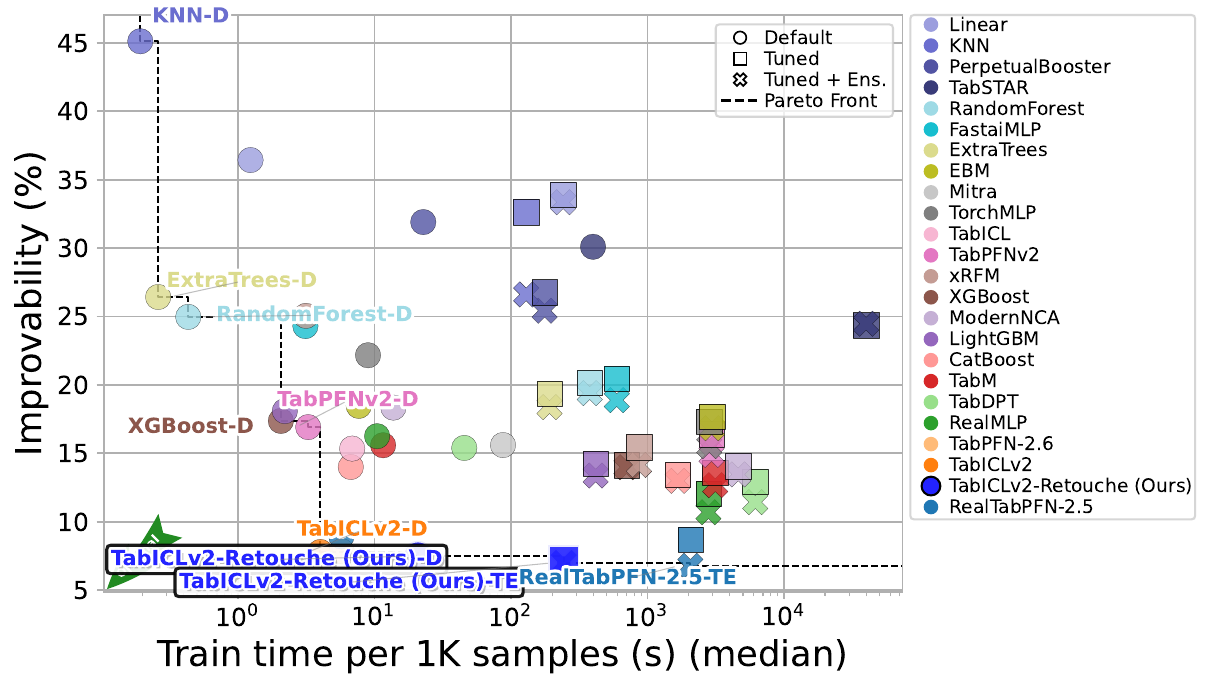}
  \caption{Improvability vs.\ training time per $1$K samples.}
  \label{fig:pareto_train}
\end{subfigure}\hfill
\begin{subfigure}{0.49\linewidth}
  \centering
  \includegraphics[width=\linewidth]{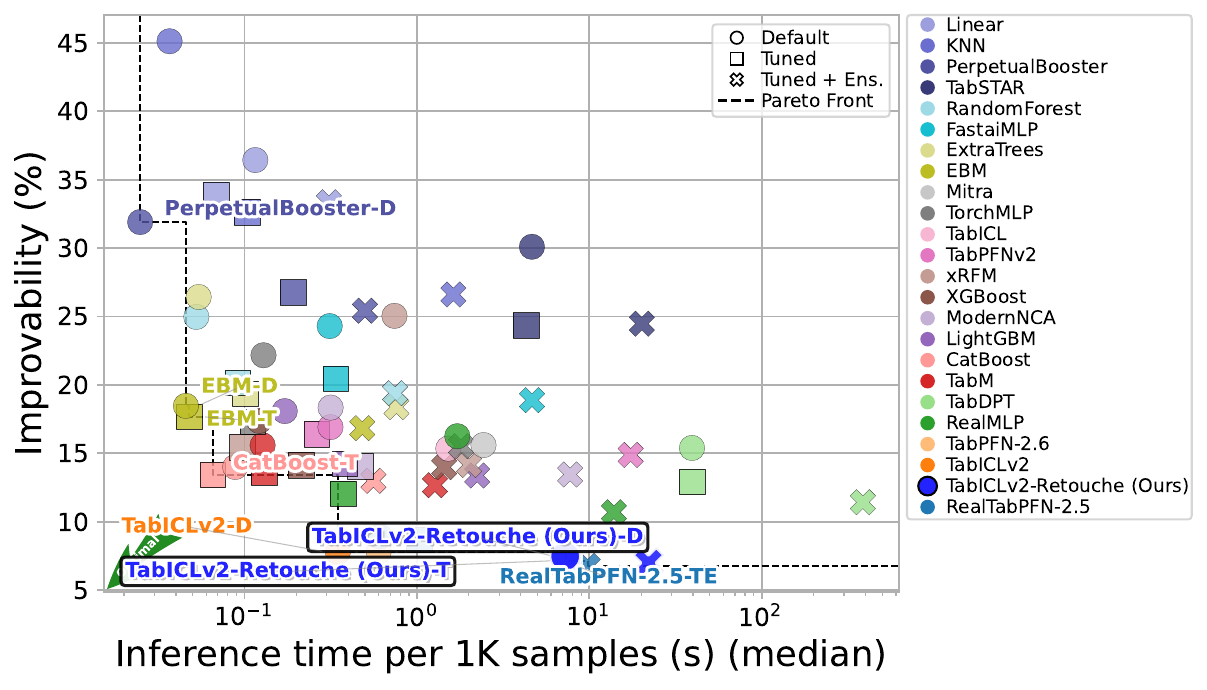}
  \caption{Improvability vs.\ inference time per $1$K samples.}
  \label{fig:pareto_infer}
\end{subfigure}
\caption{Quality vs.\ compute trade-off on TabArena-Lite: improvability vs.\ training time (a) and improvability vs.\ inference time (b). Lower-left is better; markers denote leaderboard methods.}
\label{fig:pareto}
\end{figure}

\paragraph{Improvability vs.\ compute.}
Figure~\ref{fig:pareto} replots the TabArena-Lite leaderboard data of Figure~\ref{fig:pareto_elo} with \emph{improvability} on the y-axis (the gap to the best per-task method, averaged across tasks, as defined by \citet{erickson_tabarena_2025}; lower is better). The Pareto frontier conclusion of Section~\ref{sec:exp:main} is invariant to this re-axis: \specmethod{} (T+E) retains its position as the highest-quality (lowest-improvability) operating point on the training-time Pareto frontier among tuned-and-ensembled methods, and remains on the inference-time frontier.

\begin{figure}[t]
\centering
\includegraphics[width=\linewidth]{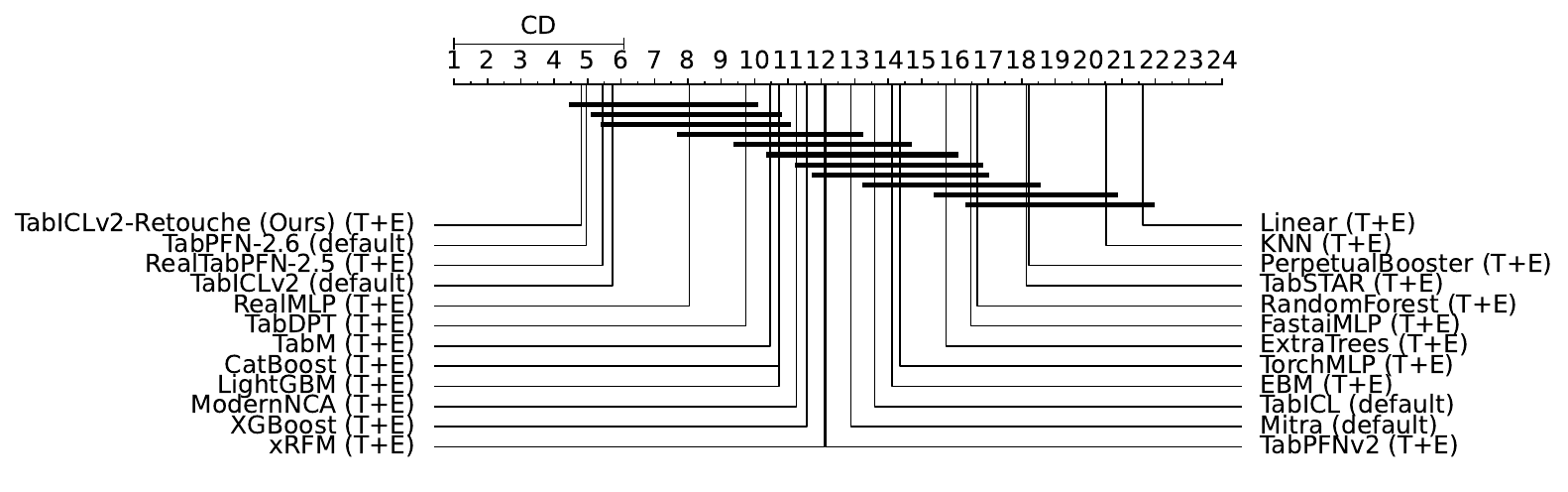}
\caption{Critical-difference diagram (Friedman test with Nemenyi post-hoc, $\alpha = 0.05$) over per-task ranks on TabArena-Lite, covering the headline \specmethod{} (cross) (T+E) run and every non-AutoGluon TabArena-Lite leaderboard method at its strongest protocol ($24$ entries total). Mean rank is on the horizontal axis (lower is better); horizontal bars connect methods whose pairwise rank difference is not statistically significant.}
\label{fig:critical-diagram}
\end{figure}

\paragraph{Critical-difference diagram.}
Figure~\ref{fig:critical-diagram} consolidates per-task ranks across TabArena-Lite into a single Demšar-style critical-difference diagram (Friedman test with Nemenyi post-hoc, $\alpha = 0.05$) over the headline \specmethod{} (cross) (T+E) run and every non-AutoGluon TabArena-Lite leaderboard method at its strongest protocol, for a total of $24$ entries. \specmethod{} (T+E) achieves the lowest mean rank, ahead of \TabPFN{}-2.6 (D), RealTabPFN-2.5 (T+E), and the unmodified \TabICL{}v2 (D). It sits in the top significance group, which spans \TabPFN{}-2.6 (D), RealTabPFN-2.5 (T+E), \TabICL{}v2 (D), RealMLP (T+E), TabDPT (T+E), and TabM (T+E); within this group the pairwise rank differences are not statistically significant at $\alpha = 0.05$, so \specmethod{}'s top position is best read as ``tied at the top'' rather than as a strict separation from the strongest baselines. The remaining tabular methods (gradient-boosted trees, the legacy TFMs, and the simpler MLP and tree baselines) form a long tail of partially overlapping significance groups extending out to KNN and Linear at the bottom of the diagram.

\section{Analysis: Where does the adapter intervene?}
\label{app:analysis}







\paragraph{Identity-guard activation rate.}
The identity guard (Section~\ref{sec:method:guard}) selects the adapted path on $67.7\%$ of runs ($3{,}040/4{,}488$, counted at the dataset $\times$ config $\times$ fold level in the official cross-block batch), where the adapter improves over the unmodified \TabICL{}v2 base on the held-out AutoGluon bag-fold validation set; on the remaining $32.3\%$ ($1{,}448/4{,}488$) the guard routes back to the base, ensuring the adapter is never deployed when it would not help. Because the validation set is held out from the per-fold trainer, this is a clean estimate of how often the adapter delivers a genuine gain.

\begin{figure}[t]
\centering
\includegraphics[width=\linewidth]{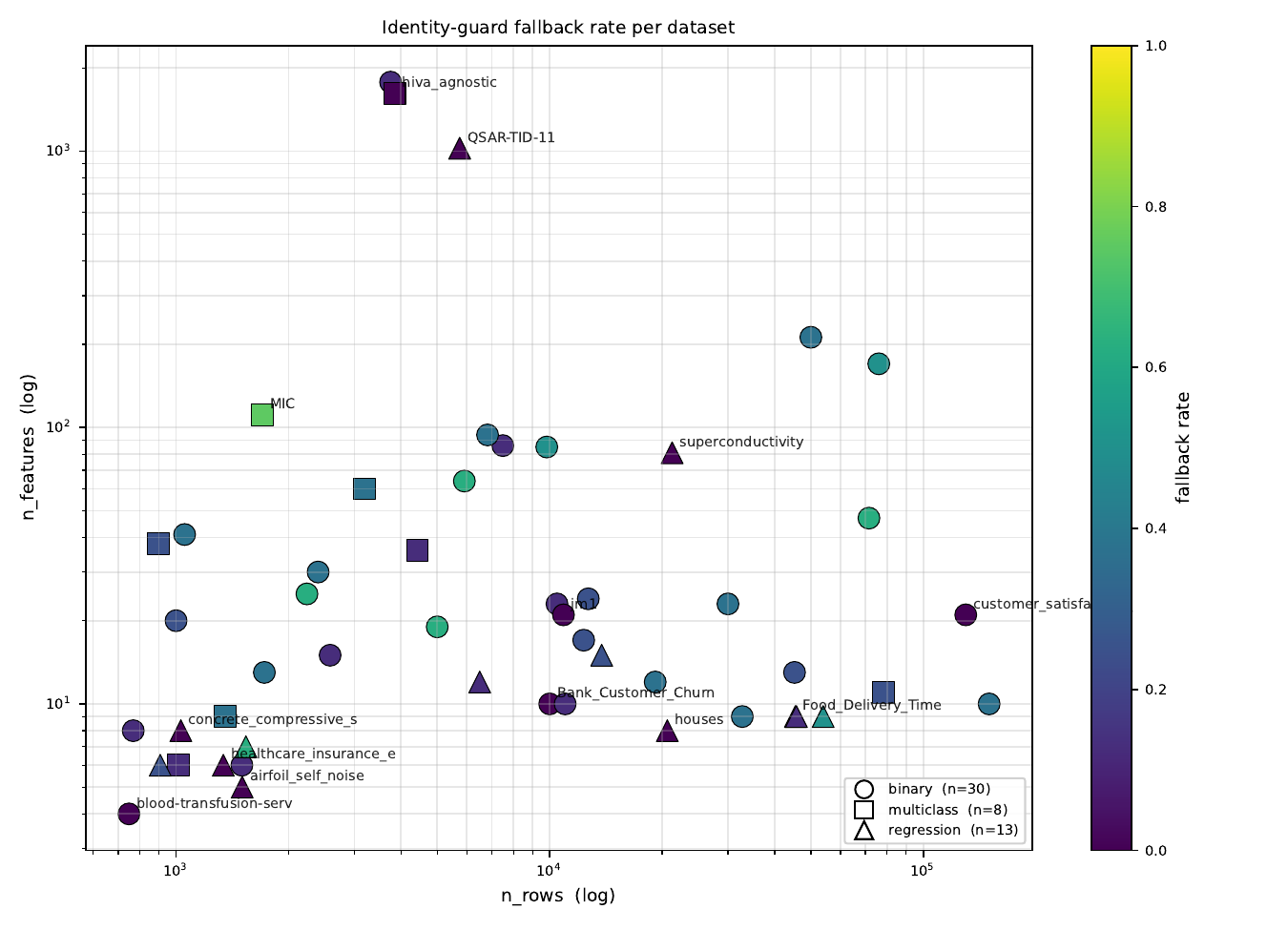}
\caption{Per-dataset identity-guard fallback rate for the headline cross-block run of \specmethod{} on TabArena-Lite. Each marker is one dataset, positioned by its number of rows (horizontal, log scale) and number of features (vertical, log scale). The color encodes the fraction of the $8$ AutoGluon bag-folds on which the guard routed back to the unmodified \TabICL{}v2 base, restricted to the single configuration that minimizes the held-out AutoGluon bag-fold validation error per dataset. Marker shape encodes problem type (circle: binary, square: multiclass, triangle: regression). Datasets at the extremes (fallback rate $\ge 0.75$ or $\le 0.05$) are annotated. 
}
\label{fig:fallback-rate-per-tid}
\end{figure}

\paragraph{Per-dataset fallback distribution.}
Figure~\ref{fig:fallback-rate-per-tid} plots the fallback rate per-dataset. For each dataset we restrict to the single configuration that minimizes the held-out AutoGluon bag-fold validation error and report the fraction of its $8$ AutoGluon bag-folds on which the guard fired. Across the $51$ dataset the mean per-dataset fallback rate is $25.2\%$ (median $25.0\%$, interquartile range $[12.5\%, 37.5\%]$), with $12/51$ datasets at $0\%$ (the trained adapter wins on every fold), no dataset at $100\%$, and a single dataset (\textsf{MIC}, multiclass with $111$ features) at the maximum of $75\%$. 

The pattern is consistent with \TabICL{}v2 being a strong starting point. Regression datasets are the most adapter-friendly (mean fallback $14.4\%$, median $0\%$ across $13$ datasets; $7/13$ at $\le 5\%$), while binary classification ($n\!=\!30$, mean $29.2\%$) and multiclass classification ($n\!=\!8$, mean $28.1\%$) show more guard activity. 

\section{Cross-block weight inspection on a financial dataset}
\label{app:weight-inspection}

The cross block in \specmethod{} computes
\begin{equation*}
x_{l+1} = x_0 \odot (W_l\, x_l + b_l) + x_l,
\end{equation*}
where $\odot$ denotes the Hadamard product. Component $i$ of the layer-$(l{+}1)$ output is a sum of terms $x_0[i]\cdot W_l[i,j]\cdot x_l[j]$, so $W_l[i,j]$ is the explicit coefficient of the multiplicative interaction $x_0[i]\cdot x_l[j]$. After fitting we can read these weights off and ask which feature pairs the adapter places weight on. 


To estimate cumulative effects across $L$ layers, we model the cross block as a vector-valued map $\mathrm{cross} : \mathbb{R}^d \to \mathbb{R}^d$, and we collapse it to the scalar function $f(x) = \sum_k \mathrm{cross}(x)_k$ by summing its $d$ output channels with equal weight, so that the second derivative of $f$ is a single $d \times d$ matrix. We then read off the symmetrised numerical Hessian $H[i,j] = \tfrac{1}{2}\big(\partial^2 f/\partial x_i\,\partial x_j + \partial^2 f/\partial x_j\,\partial x_i\big)$ via autograd, evaluated at the test-set column mean (the per-feature average of the preprocessed test inputs). Each entry $H[i,j]$ aggregates the multiplicative coupling between features $i$ and $j$ across all cross layers and all output channels, including BatchNorm rescaling and the inner activation between the low-rank factors $V_l\,U_l$ when one is present. We work with $H$ rather than the analytical aggregate $\sum_l\!\tfrac{1}{2}(W_l + W_l^\top)$ because this dataset's inner block can be low-rank with a ReLU between $V_l$ and $U_l$.

\begin{figure}[t]
\centering
\includegraphics[width=\linewidth]{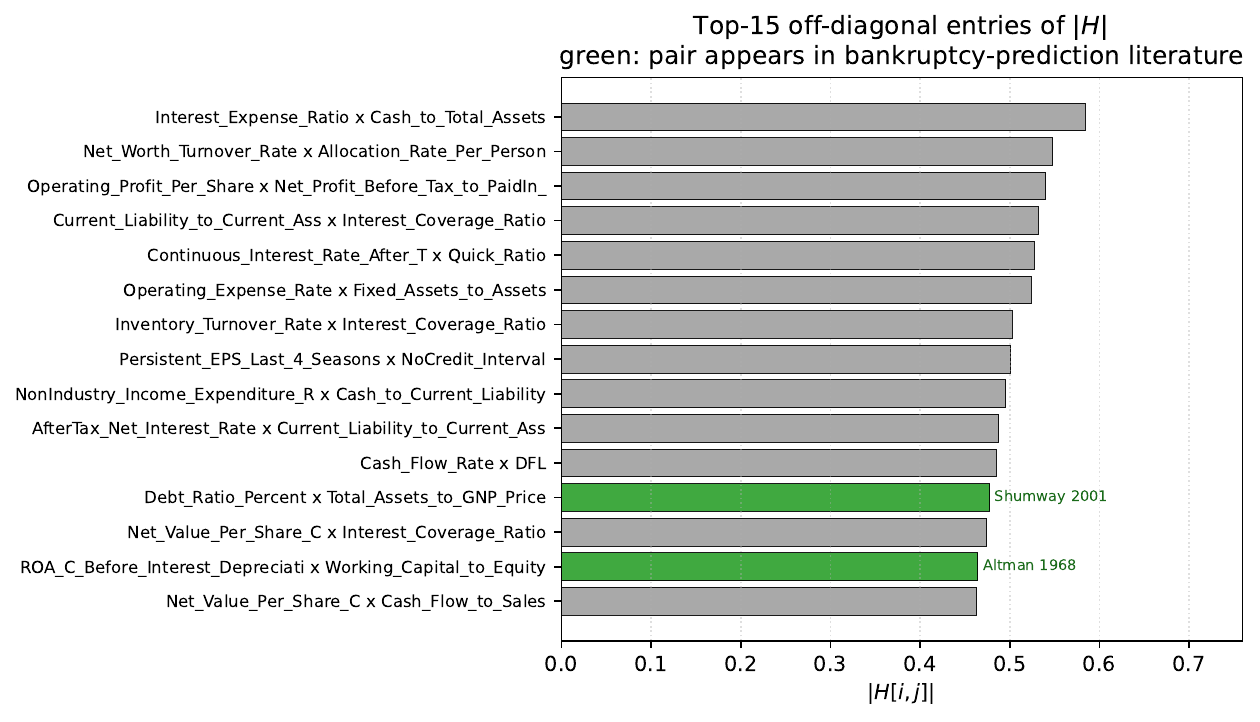}
\caption{Magnitudes of the $15$ largest off-diagonal entries of $|H|$ for the first bag-member of \specmethod{} (T) on \textsf{taiwanese\_bankruptcy\_prediction} (TabArena-Lite ID $363706$), sorted with the largest at the top. Pairs whose two ratios both appear as separate regressors in a named bankruptcy-prediction study \citep{altman_financial_1968,shumway_forecasting_2001,chava_bankruptcy_2004,ohlson_financial_1980} are coloured green; the remaining pairs are grey.}
\label{fig:weight-inspection}
\end{figure}

We analyze $|H|$ for the first bag-member of the (T) version of \specmethod{} (the per-dataset best-tuned configuration; see Section~\ref{sec:exp:setup}) on \textsf{taiwanese\_bankruptcy\_prediction} (TabArena-Lite, ID $363706$, binary classification, $94$ numeric financial-ratio features). Among the $15$ largest off-diagonal entries of $|H|$ (Figure~\ref{fig:weight-inspection}), $2$ involve feature pairs whose two ratios have both appeared as separate regressors in named bankruptcy-prediction studies.\footnote{Explicit interaction terms have been studied in this literature: \citet{laitinen_bankruptcy_2000} add second-order and pairwise products of $\mathrm{Cash}/TA$, $\mathrm{CashFlow}/TA$, and $\mathrm{Equity}/TA$ to a logit model and report improved accuracy 1--2 years before bankruptcy. The specific pairs probed there differ from those highlighted by the cross block here.} We list those alignments below without claiming the network has identified the underlying generative process:
\begin{itemize}
    \item \textsf{Debt\_Ratio\_Percent} $\times$ \textsf{Total\_Assets\_to\_GNP\_Price}. Leverage interacted with a real-deflated firm-size measure. Both ratios appear directly as regressors in the O-score logit model of \citet{ohlson_financial_1980}. Size and leverage also enter the discrete-time hazard models of \citet{shumway_forecasting_2001} and \citet{chava_bankruptcy_2004} as separate regressors.
    \item \textsf{ROA\_C\_Before\_Interest\_Depreciation} $\times$ \textsf{Working\_Capital\_to\_Equity}. A profitability and working-capital pair within the family of variables used by the discriminant Z-score of \citet{altman_financial_1968}.
\end{itemize}
The remaining top-$15$ entries involve narrower accounting variables, for example \textsf{Allocation\_Rate\_Per\_Person} or \textsf{Persistent\_EPS\_Last\_4\_Seasons}, for which we do not have a specific reference and which we report without further interpretation.

\textbf{Caveats}:
\begin{itemize}
    \item The dataset has many near-duplicate ratios (six profitability variants, five leverage variants, three liquidity variants), and the cross block carries low rank by construction. Under such multicollinearity, a low-rank fit is expected to select one proxy per economic concept rather than all of them: the textbook pair \textsf{Quick\_Ratio} $\times$ \textsf{Interest\_Expense\_Ratio} (a liquidity-versus-debt-burden interaction) appears here with small magnitude, while the closely related \textsf{Cash\_to\_Total\_Assets} $\times$ \textsf{Interest\_Expense\_Ratio} appears among the largest entries of $|H|$. The bullet list above should therefore not be read as evidence that the network has identified those specific documented variables. Substituting alternative proxies under multicollinearity is expected behavior for any low-rank fit and does not bear on which proxy is causally preferable.
    \item This is a single illustrative dataset with named features. \textbf{We do not extrapolate} the alignments observed here to other datasets in TabArena-Lite, several of which use anonymized feature names that preclude domain interpretation. The intent of this appendix is narrow: the cross block's weights are inspectable, and on this dataset $2$ of the $15$ largest entries of $|H|$ correspond to feature pairs whose two ratios both appear as separate regressors in named bankruptcy-prediction studies.
\end{itemize}

\section{Ablations}
\label{app:ablations}

We conducted ablation studies to understand the impact of each element in the design of \method{}.

\textbf{Block type (cross vs.\ MLP vs.\ identity)}: We compare the DCNv2 cross block (our default), the residual-bottleneck MLP block, and identity (no adapter, equivalent to the frozen \TabICL{}v2 backbone) at matched seeds and an otherwise identical configuration grid---the same $11$ hyperparameter draws used in the main results (Appendix~\ref{app:hpo}). The cross-block run is the headline batch reported in Section~\ref{sec:exp:main}; the paired MLP-block reuses every other knob unchanged.

\textbf{Training ablation (Random Adapter)}: We observed that the identity guard (Section~\ref{sec:method:guard}) routed back to the unmodified frozen \TabICL{}v2 on a sizable fraction of runs: $32.3\%$ at the run level ($1{,}448/4{,}488$ = 51 datasets x 11 configurations x 8 folds) in the headline cross-block batch, where the trained adapter did not improve over the base on the held-out fold. This raised a natural concern: does the lift come from end-to-end training of the adapter, or could it instead be explained by randomness and the $8$-fold ensembling? To isolate the contribution of training, we re-run \specmethod{} with the adapter randomly initialized at the beginning of the training and never updated. All the other settings are all held identical to the headline run; only the gradient updates to the adapter are removed.



\textbf{Gate and identity-guard variants}: The block-type and Random Adapter ablations probe two of the three safety knobs of \specmethod{}: the choice of inner block $\delta$ and end-to-end optimization of the adapter parameters. We complement them with four further runs that probe the trainable per-channel gate $\alpha$ (Section~\ref{sec:method:adapter}) and the identity guard (Section~\ref{sec:method:guard}). 

All the \specmethod{} variants share the headline $11$-configurations and the $8$-fold AutoGluon bagging protocol; each variant modifies a single knob:
\begin{itemize}
    \item \textsf{[cross]} Headline cross-block run.
    \item \textsf{[mlp]} Block-type ablation: residual-bottleneck MLP $\delta$ in place of the DCNv2 cross block.
    \item \textsf{[cross, alpha\_init$+0.5$]} Gate-init ablation: per-config $\alpha$ initialization shifted by $+0.5$ from the headline default, $\alpha$ remains trainable.
    \item \textsf{[cross, alpha$=1$]} Strict-freeze ablation: $\alpha$ initialized at $1$ and held fixed; the adapter reduces to $\frozenTFM \circ \delta$ with no gated residual.
    \item \textsf{TabICLv2} The unmodified frozen \TabICL{}v2 base run at its default (D) configuration: no adapter, no gate, no identity guard. Serves as the no-adapter floor against which each \specmethod{} variant is measured.
    \item \textsf{[cross, random]} Training ablation (Random Adapter, above): $\delta$ randomly initialized and never updated.
    \item \textsf{[cross, no-guard]} Guard ablation: identity guard disabled, so the trained adapter is always deployed.
    \item \textsf{[cross, alpha$=1$, no-guard]} Combined ablation: $\alpha$ frozen at $1$ and guard disabled.
\end{itemize}

\begin{figure}[t]
\centering
\includegraphics[width=\linewidth]{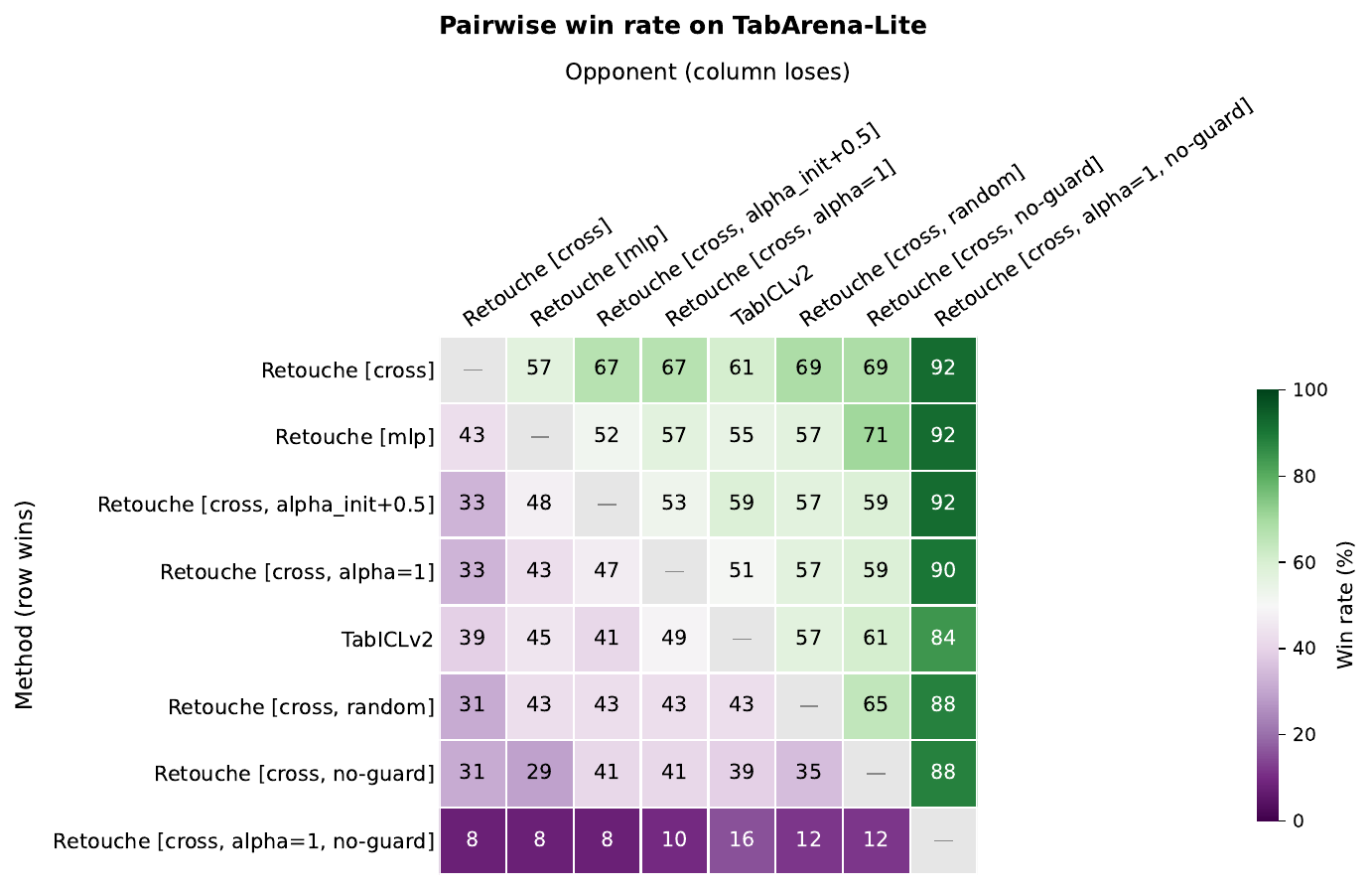}
\caption{Pairwise win rates (\%) among \specmethod{} variants (T+E) and the unmodified \TabICL{}v2 (D) on TabArena-Lite. Cell $(i,j)$ reports the percentage of datasets on which row method $i$ outperforms column method $j$.}
\label{fig:winrate-matrix}
\end{figure}

We report pairwise win rates rather than Elo here: the pool consists many variants of one method, so an Elo fit would inflate the headline rating with easy wins against its own degenerate ablations. A pairwise win rate, by contrast, is a direct head-to-head statistic that does not depend on the rest of the pool. Figure~\ref{fig:winrate-matrix} reports pairwise win rates (\%) among the \specmethod{} variants (T+E) and the unmodified \TabICL{}v2 base (D) on TabArena-Lite. Cell $(i,j)$ is the percentage of datasets on which row method $i$ outperforms column method $j$. The headline cross-block run leads the matrix, beating every other entry by at least $57$--$43$. The cross-vs-MLP edge ($57$--$43$) is the narrowest non-degenerate gap. Both gate ablations cost the same: \textsf{[cross, alpha\_init$+0.5$]} and \textsf{[cross, alpha$=1$]} each lose to the headline at $33$--$67$, indicating that the small near-identity initialization and the trainability of $\alpha$ contribute to the performance of the headline. The training ablation \textsf{[cross, random]} loses at $31$--$69$, and the guard ablation \textsf{[cross, no-guard]} loses at the same $31$--$69$. The doubly-ablated variant \textsf{[cross, alpha$=1$, no-guard]} wins only $8$--$16\%$ across all opponents. Without either safety mechanism the trained adapter actively hurts on average.

The relative performance of the variants to the unmodified \TabICL{}v2 base sharpens the picture. First, the headline cross-block lifts the base by $61$--$39$ in pairwise wins and the MLP block by $55$--$45$, so both inner blocks deliver a lift over no-adapter. Second, the strict-freeze ablation \textsf{[cross, alpha$=1$]} essentially \emph{ties} the base at $51$--$49$: with the gate frozen at $1$ and the identity guard intact, the framework is no better than running \TabICL{}v2 alone, so the small-init trainable $\alpha$ is what allows the lift to materialize on top of the guard. 

The most striking cell is \textsf{[cross, random]} vs.\ \textsf{[cross, no-guard]}: the untrained-but-guarded random adapter wins $65$--$35$ over the trained-but-unguarded adapter. On average across TabArena-Lite, the safety floor of the identity guard contributes more to the headline lift than end-to-end optimization of the adapter does.

The pairwise comparisons support a layered reading of \specmethod{}'s components:
\begin{itemize}
    \item The identity guard is the single most important component: it provides more pairwise wins than training the adapter end-to-end (\textsf{[cross, random]} beats \textsf{[cross, no-guard]} $65$--$35$), and removing it drives the trained adapter \emph{below} the unmodified \TabICL{}v2 base ($39$--$61$).
    \item The trainable gated residual is the second-most important: with the guard intact, freezing $\alpha$ at $1$ erases the framework lift and only \emph{ties} the base ($51$--$49$), and shifting the initialization by $+0.5$ keeps a $59$--$41$ edge over the base while losing $33$--$67$ to the headline default.
    \item Removing both safety mechanisms simultaneously is catastrophic: \textsf{[cross, alpha$=1$, no-guard]} wins only $8$--$16\%$ across all opponents and loses to the unmodified base $16$--$84$, validating the framework prescription that the gated residual and the identity guard are jointly load-bearing rather than redundant safeguards.
\end{itemize}

\section{Head-to-head: external adaptation baselines}
\label{app:h2h}

The Ablations appendix (Appendix~\ref{app:ablations}) probes \specmethod{}'s internal design choices. Here we instead compare the framework against the three contemporary adaptation paradigms applicable to a frozen \TabICL{}v2 stack: weight-space full supervised fine-tuning, PEFT (LoRA), and input-space encoder-and-bagging (BETA). 

All four trained methods share the headline $11$-configurations HPO grid (Appendix~\ref{app:hpo}) and the $8$-fold AutoGluon bagging protocol; each baseline is run via the public reference implementation indicated below:
\begin{itemize}
    \item \textsf{TabICLv2-Retouche}: The headline cross-block run reported in Section~\ref{sec:exp:main}.
    \item \textsf{TabICLv2}: Unmodified frozen \TabICL{}v2~\citep{qu_tabiclv2_2026} at its default configuration.
    \item \textsf{TabICLv2-LoRA (TabTune)}: we vendor TabTune~\citep{tanna_tabtune_2025}, reusing only its \texttt{apply\_tabular\_lora} helper to inject LoRA adapters into the \TabICL{}v2 stack. The surrounding training loop, optimizer setup, batching, ICL context/query split, early stopping, and identity guard are the iso-Retouche pipeline used by the headline batch (we do not invoke TabTune's own \texttt{simple\_sft} or \texttt{meta-learning} fine-tuning entry points). All backbone parameters are explicitly frozen before injection so that only the LoRA \texttt{lora\_A}/\texttt{lora\_B} weights are trainable.
    \item \textsf{TabICLv2-SFT (TabTune)}: every backbone parameter is unfrozen and updated end-to-end. The TabTune library is not invoked here; we run the iso-Retouche training loop on the unfrozen backbone with the SFT-style hyperparameters reported in TabTune's reference recipe~\citep{tanna_tabtune_2025} (Adam, learning rate $10^{-5}$, cross-entropy). The ``(TabTune)'' tag therefore identifies the recipe, not the executed code path: the per-dataset HPO grid, the bagging protocol, and the identity guard are the same as for the LoRA arm and the headline batch.
    \item \textsf{BETA-TabICLv2}: the BETA recipe~\citep{liu_tabpfn_2025} (a two-layer fixed-width MLP encoder placed before the frozen TFM and trained end-to-end) ported to the \TabICL{}v2 backbone. The published BETA implementation targets first-generation \TabPFN{}; we re-host its encoder/training loop on top of \TabICL{}v2 so the head-to-head is iso-backbone with the other rows. Our wrapper currently runs single-forward inference rather than the published $16$-encoder bootstrap bagging at predict time, so the BETA-TabICLv2 numbers reported here should be read as a lower bound on what the published recipe would deliver on \TabICL{}v2.
\end{itemize}

\begin{figure}[t]
\centering
\includegraphics[width=0.78\linewidth]{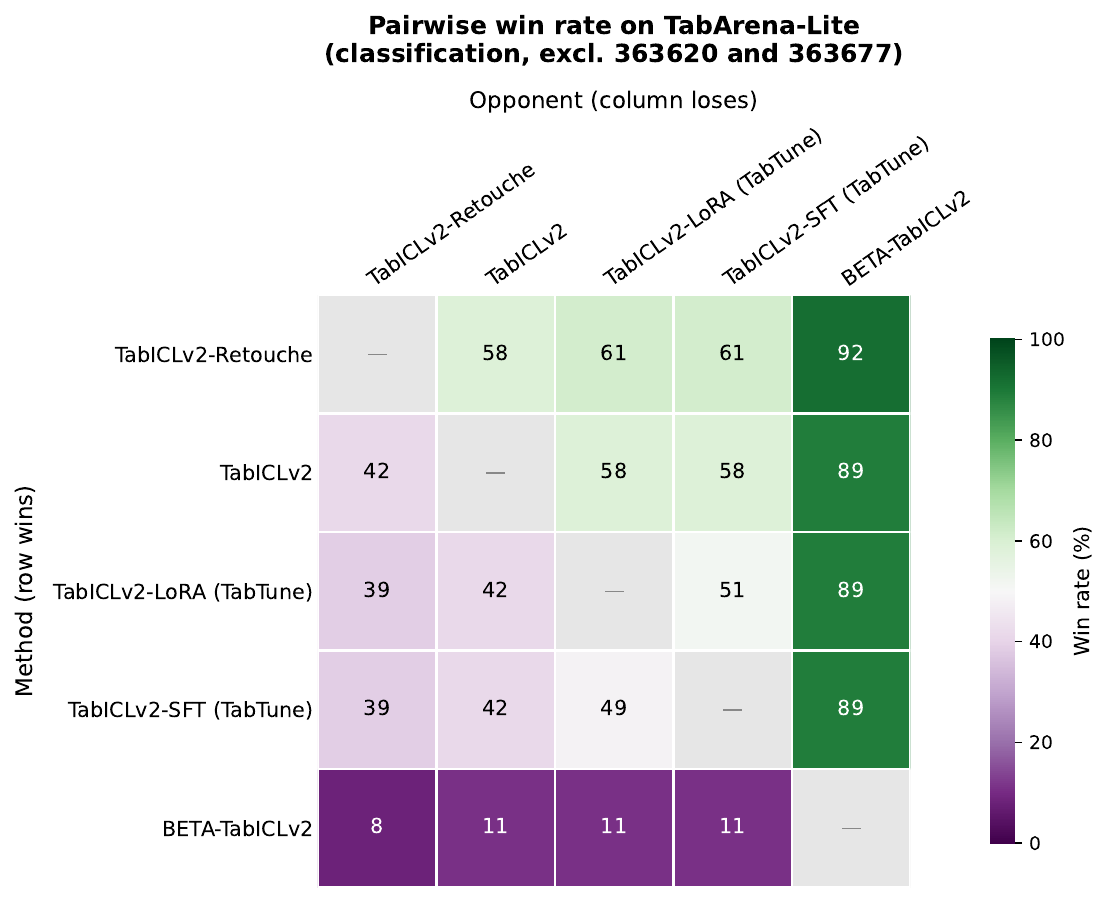}
\caption{Pairwise win rates (\%) for the four iso-backbone adaptation paradigms and the unmodified base on the TabArena-Lite classification subset ($36$ datasets, two datasets excluded for baseline OOM). 
}
\label{fig:winrate-matrix-h2h}
\end{figure}

The matrix in Figure~\ref{fig:winrate-matrix-h2h} reports pairwise win rates (\%) on the TabArena-Lite classification subset (\citet{erickson_tabarena_2025}'s $38$-task classification block, minus two tasks: Bioresponse (363620) and hiva\_agnostic (363677), both excluded because the gradient-path baselines---LoRA and full SFT---could not complete on them within available GPU memory (A100); the comparison scope is therefore $36$ datasets. \specmethod{} itself completes on all $38$ classification tasks. The exclusion here is solely to keep the head-to-head matrix iso-protocol across all four trained methods. 

The matrix supports four findings.
\begin{itemize}
    \item \specmethod{} is the only adaptation strategy that beats the unmodified \TabICL{}v2 base. It wins $58$--$42$ over the base, $61$--$39$ over both LoRA and SFT, and $92$--$8$ over BETA. Among the five rows it is the only one with strictly above-$50\%$ win rates against every other entry in the comparison pool.
    \item Weight-space adaptation (both LoRA and full SFT) does not improve over zero-shot \TabICL{}v2 on this subset. LoRA loses to the base $42$--$58$ and full SFT loses by the same $42$--$58$. With our matched $11$-configuration HPO budget and TabArena-Lite's per-task scale, neither parameter-efficient nor full weight-space tuning yields a measurable gain over the frozen backbone, consistent with the cross-model finding of \citet{tanna_exploring_2026} that weight-space gains on already-strong zero-shot TFMs are model- and data-dependent.
    \item LoRA and full SFT are essentially indistinguishable on this subset ($51$--$49$ in either direction). Holding the backbone, the HPO grid, and the bagging protocol fixed, the two weight-space recipes converge to the same per-task error distribution, so the choice between PEFT and full fine-tuning is dominated by their compute cost (full SFT carries the optimizer state for every backbone parameter; LoRA does not) rather than by accuracy.
    \item Replacing the input rather than nudging it actively hurts when the backbone is already strong. BETA-TabICLv2 loses to the bare base $11$--$89$. The BETA encoder was designed against first-generation \TabPFN{}'s capacity constraints, where compressing the input was practically necessary to keep the in-context window tractable. Mounted on \TabICL{}v2, where that constraint no longer binds, the same fixed-width MLP encoder destroys information that the backbone could otherwise consume directly. \specmethod{}'s near-identity, dimension-preserving residual nudge avoids this failure mode by leaving the original feature representation intact and intervening only in proportion to a learnable per-channel gate.
\end{itemize}

Taken together, the matrix supports the central positioning of the framework: among the four adaptation paradigms compatible with a strong frozen TFM, the input-space residual nudge is the only one that delivers a measurable lift over the zero-shot backbone on TabArena-Lite classification. The bottleneck has indeed shifted from capacity (where BETA's encoder helped on TabPFN-v1) to alignment (where every weight-space path tested here ties or loses to the base, and only \specmethod{} clears it).

\section{Implementation details}
\label{app:impl}

\paragraph{Preprocessing.}
We map all input features to a common continuous representation before applying the adapter. Numeric features are median-imputed and standardized. Categorical features are encoded as ordinal integers and then standardized to the same scale. This choice keeps the input dimensionality fixed and ensures that all coordinates enter the adapter on a comparable scale, which is particularly important for the cross block, whose Hadamard products are sensitive to feature magnitudes.

We also consider different options for preprocessing:
\begin{itemize}
    \item Robust encoder \citep{holzmuller_better_2025} and numerical encoding, such as PLR, for numeric features \citep{gorishniy_embeddings_2023}
    \item Mixed preprocessing variant in which low-cardinality categorical features (at most $8$ levels) are one-hot encoded, while higher-cardinality features remain ordinal-encoded and standardized \citep{holzmuller_better_2025}
    \item Target embedding or learnable Categorical embedding for categorical features \citep{holzmuller_better_2025}
\end{itemize}
These alternatives can encode numerical better, and/or expose finer-grained categorical structure, but at the cost of increasing dimensionality and weakening the ``close-to-raw-input'' character of the overall pipeline. Unless otherwise stated, the experiments reported in this paper use the default variant (standard scaler for numerical features and ordinal encoding for categorical features).

\paragraph{Other.}
To keep adaptation compatible with the backbone across a wide range of input dimensionalities, we cap the feature dimension seen by the TFM at $500$. For datasets with $d > 500$, we therefore insert a fixed-rank trainable linear projection between the adapter and the backbone; this projection is orthogonally initialized and serves only to meet the backbone's effective input budget, not to alter the adapter formulation itself. We also support Truncated-SVD as a non-learned alternative. We additionally reshuffle the context/query split at each epoch. Finally, the frozen TFM forward follows the same automatic precision policy as \TabICL{}: fp32 is used on smaller problems ($n < 1024$, $d < 60$), fp16 autocast on medium and large problems, and FA3 attention on sufficiently large inputs ($n \geq 10{,}240$). These choices are purely computational and leave the adapter architecture unchanged.

\section{Hyperparameter search space}
\label{app:hpo}

For each TabArena-Lite dataset, we evaluate $11$ configurations: one default configuration plus $10$ random draws from the search space below. The same $11$ configurations are evaluated on every dataset; they are sampled once with a fixed seed and reused across the headline batch and its companion batches (random adapter, no-fallback ablation, TabTune-LoRA, TabTune-SFT, and BETA head-to-heads, etc.). Numeric parameters are drawn from continuous ranges either uniformly or log-uniformly; categorical parameters are drawn uniformly from their listed sets; integer parameters are drawn uniformly over the listed inclusive range. The default values in Table~\ref{tab:hpo} are those of configuration $c_1$ of the batch reported in Section~\ref{sec:exp:main}.


\begin{table}[t]
\caption{Hyperparameter search space for \specmethod{}. Defaults match the first of the $11$ configurations evaluated in the headline cross-block run (Section~\ref{sec:exp:main}).}
\label{tab:hpo}
\centering
\small
\begin{tabular}{lllc}
\toprule
\textbf{Hyperparameter} & \textbf{Distribution} & \textbf{Range / choices} & \textbf{Default} \\
\midrule
\multicolumn{4}{l}{\textit{Adapter architecture}} \\
\texttt{num\_layers}     & uniform choice    & $\{1, 2\}$              & $2$ \\
\texttt{low\_rank\_ratio}$^{a}$ & uniform / full-rank & $[0.1, 0.5] \cup \{\text{None}\}$ & $0.25$ \\
\texttt{hidden\_dim}     & fixed             & $64$                    & $64$ \\
\texttt{use\_batch\_norm}& uniform choice    & $\{\text{False}, \text{True}\}$ & True \\
\texttt{alpha\_init}     & log-uniform       & $[0.01, 0.1]$           & $0.02$ \\
\texttt{alpha\_shape}    & uniform choice    & $\{\text{per-channel}, \text{global}\}$ & per-channel \\
\texttt{gate\_lr\_factor}& log-uniform       & $[2.0, 10.0]$           & $3.0$ \\
\texttt{block\_type}$^{b}$& fixed (cross-arm) & \texttt{cross}          & cross \\
\midrule
\multicolumn{4}{l}{\textit{Optimization}} \\
\texttt{optimizer}       & uniform choice    & $\{\text{AdamW}, \text{Muon}\}$ & AdamW \\
\texttt{lr}              & log-uniform       & $[10^{-3},\, 1.5\!\times\!10^{-2}]$ & $5\!\times\!10^{-3}$ \\
\texttt{weight\_decay}   & log-uniform       & $[10^{-3},\, 5\!\times\!10^{-2}]$   & $3\!\times\!10^{-3}$ \\
\texttt{max\_grad\_norm} & log-uniform       & $[1.0, 5.0]$            & $2.0$ \\
\texttt{label\_smoothing}& uniform           & $[0.05, 0.30]$          & $0.15$ \\
\texttt{beta2}           & uniform           & $[0.95, 0.99]$          & $0.97$ \\
\midrule
\multicolumn{4}{l}{\textit{Training schedule}} \\
\texttt{epochs}          & integer uniform   & $\{100, \ldots, 200\}$  & $150$ \\
\texttt{patience}        & integer uniform   & $\{10, \ldots, 15\}$    & $10$ \\
\texttt{lr\_schedule}    & uniform choice    & $\{\text{cosine}, \text{coslog4}\}$ & coslog4 \\
\midrule
\multicolumn{4}{l}{\textit{Initialization and preprocessing}} \\
\texttt{weight\_init}    & uniform choice    & $\{\text{xavier-normal}, \text{small-normal}\}$ & small-normal \\
\texttt{activation}      & uniform choice    & $\{\text{None}, \text{ReLU}\}$ & None \\
\texttt{preprocessor}    & uniform choice    & $\{\text{ordinal-scaled}, \text{onehot-ordinal}\}$ & ordinal-scaled \\
\bottomrule
\end{tabular}\\[4pt]
{\footnotesize $^{a}$\texttt{low\_rank\_ratio} applies only when \texttt{block\_type} is \texttt{cross}; sampled as full rank (\texttt{None}) with probability $1/3$ and otherwise uniformly from $[0.1, 0.5]$.\quad $^{b}$ The result reported in Section~\ref{sec:exp:main} has \texttt{block\_type} fixed to \texttt{cross}; the paired $\{\text{cross}, \text{mlp}\}$ sweep is the block-type ablation in Appendix~\ref{app:ablations}.}
\end{table}

\section{Computational details}
\label{app:cost}

\paragraph{Compute platform.} All \specmethod{} runs reported in this paper were submitted as Databricks job runs on Microsoft Azure. Two Azure GPU SKUs were used:
\begin{itemize}
\item \textbf{A10:} \texttt{Standard\_NV36ads\_A10\_v5} ($1{\times}$ NVIDIA A10, $24$\,GB HBM).
\item \textbf{A100:} \texttt{Standard\_NC24ads\_A100\_v4} ($1{\times}$ NVIDIA A100, $80$\,GB HBM).
\end{itemize}

\paragraph{Sharding.} The headline batch reported in Section~\ref{sec:exp:main} was split across $17$ Databricks job runs: $6$ A10 job clusters each handling a $1/6$ shard of the lighter TabArena-Lite tasks via the official sharded suite, plus $11$ A100 job clusters each pinned to a single heavier task (task IDs $363616$, $363620$, $363628$, $363630$, $363631$, $363673$, $363677$, $363683$, $363697$, $363699$, $363705$). Each shard ran the $11$ HPO configurations (one default plus 10 random draws; full search space in Appendix~\ref{app:hpo}) over the standard $8$-fold AutoGluon bagging protocol~\citep{erickson_tabarena_2025}, for a total of $51 \times 11 \times 8 = 4{,}488$ end-to-end fits per batch.

\paragraph{Hardware-asymmetry caveat.} Time-per-$1$K-samples figures reported on the Pareto plots (Figures~\ref{fig:pareto_elo} and \ref{fig:pareto}) were measured on this A10/A100 mix, while published TabArena baselines for GPU methods were measured on faster H100s. As discussed in Section~\ref{sec:exp:main}, this biases the comparison \emph{against} \specmethod{}; under matched hardware its wall-clock numbers would only shift further toward the cheap-time end of the plots.

\todo{Wall-clock breakdown on representative dataset sizes; GPU memory profile of the grad-enabled \TabICL{} forward.}

\section{Full TabArena-Lite leaderboard}
\label{app:leaderboard}

For completeness, Table~\ref{tab:full-leaderboard} reports the complete TabArena-Lite leaderboard for the headline cross-block run of \specmethod{} (Section~\ref{sec:exp:main}), covering all baseline methods at all three evaluation protocols: (D) default configuration, (T) tuned over $10$ random HPO configurations, and (T+E) tuned and ensembled across the $8$ AutoGluon bagging folds. Methods are ranked by Elo with bootstrap $95\%$ confidence intervals shown as ${}_{-l,+u}$ subscripts. Within each column, the best entry is rendered in \textbf{bold} and the second best in \textit{italic}. Train and predict times are reported per $1{,}000$ samples; as discussed in Appendix~\ref{app:cost}, our runs use a mix of A10 and A100 GPUs while published baselines were measured on H100, which biases the time comparison \emph{against} \specmethod{}.

{\footnotesize
\setlength{\tabcolsep}{3pt}
\begin{longtable}{llcccccrr}
\caption{Full TabArena-Lite leaderboard for the headline cross-block run of \specmethod{} (Section~\ref{sec:exp:main}). Ranked by Elo. Best per column in \textbf{bold}, second in \textit{italic}. Train/predict times are per $1{,}000$ samples.}
\label{tab:full-leaderboard}\\
\toprule
\textbf{Model} & \textbf{Elo ($\uparrow$)} & \textbf{Norm.} & \textbf{Avg.} & \textbf{Harm.} & \textbf{\#wins} & \textbf{Improva-} & \textbf{Train time} & \textbf{Predict time} \\
 &  & \textbf{score ($\uparrow$)} & \textbf{rank ($\downarrow$)} & \textbf{mean} & \textbf{($\uparrow$)} & \textbf{bility ($\downarrow$)} & \textbf{per 1K [s]} & \textbf{per 1K [s]} \\
 &  &  &  & \textbf{rank ($\downarrow$)} &  &  &  &  \\
\midrule
\endfirsthead
\multicolumn{9}{l}{\itshape (continued)}\\
\toprule
\textbf{Model} & \textbf{Elo ($\uparrow$)} & \textbf{Norm.} & \textbf{Avg.} & \textbf{Harm.} & \textbf{\#wins} & \textbf{Improva-} & \textbf{Train time} & \textbf{Predict time} \\
 &  & \textbf{score ($\uparrow$)} & \textbf{rank ($\downarrow$)} & \textbf{mean} & \textbf{($\uparrow$)} & \textbf{bility ($\downarrow$)} & \textbf{per 1K [s]} & \textbf{per 1K [s]} \\
 &  &  &  & \textbf{rank ($\downarrow$)} &  &  &  &  \\
\midrule
\endhead
\midrule
\multicolumn{9}{r}{\itshape (continues on next page)}\\
\endfoot
\bottomrule
\endlastfoot
AutoGluon 1.5 (extreme, 4h) & \textbf{1662${}_{-84,+105}$} & \textbf{0.710} & \textbf{9.8} & \textbf{3.2} & \textbf{9.0} & \textbf{5.1\%} & 293.65 & 4.36 \\
\specmethod{} (Ours) (T+E) & \textit{1651${}_{-63,+85}$} & \textit{0.665} & \textit{10.2} & 5.0 & 1.3 & 7.0\% & 243.17 & 22.03 \\
\specmethod{} (Ours) (T) & 1632${}_{-61,+74}$ & 0.659 & 11.0 & 4.3 & 5.3 & 7.1\% & 243.17 & 7.33 \\
\TabPFN{}-2.6 (D) & 1627${}_{-61,+82}$ & 0.643 & 11.3 & 5.3 & 3.0 & 7.8\% & 5.75 & 0.60 \\
RealTabPFN-2.5 (T+E) & 1614${}_{-66,+88}$ & 0.630 & 11.8 & 4.2 & 4.0 & \textit{6.8\%} & 2059.94 & 9.79 \\
\specmethod{} (Ours) (D) & 1608${}_{-62,+67}$ & 0.623 & 12.1 & 5.9 & 1.3 & 7.5\% & 20.80 & 7.24 \\
\TabICL{}v2 (D) & 1595${}_{-67,+89}$ & 0.633 & 12.7 & \textit{4.0} & \textit{6.0} & 7.8\% & 4.01 & 0.35 \\
AutoGluon 1.4 (extreme, 4h) & 1583${}_{-68,+82}$ & 0.605 & 13.3 & 5.8 & 2.0 & 8.2\% & 556.15 & 6.31 \\
RealTabPFN-2.5 (T) & 1561${}_{-64,+74}$ & 0.556 & 14.3 & 6.9 & 2.0 & 8.7\% & 2059.94 & 1.03 \\
RealTabPFN-2.5 (D) & 1529${}_{-47,+66}$ & 0.508 & 16.0 & 9.0 & 0.0 & 8.9\% & 5.71 & 0.61 \\
AutoGluon 1.4 (best, 4h) & 1528${}_{-59,+56}$ & 0.502 & 16.1 & 7.7 & 1.0 & 9.7\% & 1754.94 & 1.77 \\
RealMLP (T+E) & 1500${}_{-44,+54}$ & 0.458 & 17.7 & 10.6 & 1.0 & 10.7\% & 2791.97 & 13.89 \\
TabDPT (T+E) & 1430${}_{-59,+69}$ & 0.403 & 22.0 & 7.4 & 3.0 & 11.4\% & 6154.73 & 386.17 \\
RealMLP (T) & 1428${}_{-59,+52}$ & 0.378 & 22.0 & 12.8 & 0.0 & 12.0\% & 2791.97 & 0.37 \\
LightGBM (T+E) & 1400${}_{-41,+41}$ & 0.275 & 24.0 & 18.9 & 0.0 & 13.4\% & 416.56 & 2.24 \\
TabM (T+E) & 1399${}_{-43,+64}$ & 0.333 & 24.0 & 15.4 & 0.0 & 12.7\% & 3133.91 & 1.27 \\
TabDPT (T) & 1388${}_{-57,+75}$ & 0.355 & 24.7 & 10.4 & 0.0 & 12.9\% & 6154.73 & 39.45 \\
CatBoost (T+E) & 1388${}_{-45,+60}$ & 0.291 & 24.7 & 16.9 & 0.0 & 13.0\% & 1665.53 & 0.56 \\
ModernNCA (T+E) & 1371${}_{-59,+76}$ & 0.343 & 25.9 & 13.1 & 1.0 & 13.5\% & 4618.50 & 7.74 \\
CatBoost (T) & 1364${}_{-48,+47}$ & 0.265 & 26.4 & 17.6 & 0.0 & 13.4\% & 1665.53 & 0.07 \\
XGBoost (T+E) & 1360${}_{-46,+43}$ & 0.231 & 26.7 & 19.2 & 0.0 & 14.0\% & 700.96 & 1.44 \\
CatBoost (D) & 1343${}_{-47,+42}$ & 0.229 & 27.8 & 18.5 & 0.0 & 14.0\% & 6.70 & 0.09 \\
LightGBM (T) & 1342${}_{-47,+50}$ & 0.221 & 27.9 & 22.9 & 0.0 & 14.3\% & 416.56 & 0.38 \\
ModernNCA (T) & 1341${}_{-57,+62}$ & 0.263 & 28.0 & 14.7 & 1.0 & 14.1\% & 4618.50 & 0.47 \\
TabM (T) & 1339${}_{-56,+62}$ & 0.261 & 28.1 & 17.9 & 0.0 & 13.6\% & 3133.91 & 0.13 \\
XGBoost (T) & 1339${}_{-49,+46}$ & 0.207 & 28.1 & 20.0 & 0.0 & 14.1\% & 700.96 & 0.21 \\
xRFM (T+E) & 1334${}_{-49,+60}$ & 0.252 & 28.5 & 18.0 & 0.0 & 14.1\% & 866.11 & 2.01 \\
\TabPFN{}v2 (T+E) & 1328${}_{-72,+75}$ & 0.312 & 28.9 & 11.4 & 1.0 & 14.9\% & 2942.08 & 17.37 \\
Mitra (D) & 1301${}_{-70,+56}$ & 0.252 & 30.8 & 14.5 & 1.0 & 15.6\% & 87.34 & 2.43 \\
xRFM (T) & 1290${}_{-45,+56}$ & 0.184 & 31.6 & 16.5 & 1.0 & 15.5\% & 866.11 & 0.10 \\
TabDPT (D) & 1284${}_{-70,+72}$ & 0.253 & 32.0 & 16.4 & 0.0 & 15.4\% & 45.42 & 39.41 \\
TabM (D) & 1278${}_{-51,+50}$ & 0.199 & 32.5 & 24.1 & 0.0 & 15.6\% & 11.56 & 0.13 \\
\TabICL{} (D) & 1278${}_{-60,+59}$ & 0.223 & 32.5 & 13.4 & 1.0 & 15.3\% & 6.86 & 1.52 \\
EBM (T+E) & 1264${}_{-52,+50}$ & 0.156 & 33.4 & 24.3 & 0.0 & 16.8\% & 2961.52 & 0.48 \\
\TabPFN{}v2 (T) & 1260${}_{-62,+78}$ & 0.214 & 33.8 & 19.8 & 0.0 & 16.4\% & 2942.08 & 0.26 \\
RealMLP (D) & 1253${}_{-50,+43}$ & 0.116 & 34.2 & 26.2 & 0.0 & 16.2\% & 10.44 & \textit{1.71} \\
TorchMLP (T+E) & 1253${}_{-54,+46}$ & 0.138 & 34.2 & 28.6 & 0.0 & 15.5\% & 2832.80 & 1.80 \\
\TabPFN{}v2 (D) & 1228${}_{-79,+66}$ & 0.189 & 36.0 & 18.6 & 0.0 & 16.9\% & 3.27 & 0.32 \\
EBM (T) & 1218${}_{-57,+56}$ & 0.114 & 36.8 & 26.4 & 0.0 & 17.7\% & 2961.52 & \textbf{0.05} \\
ModernNCA (D) & 1215${}_{-49,+59}$ & 0.114 & 37.0 & 18.5 & 1.0 & 18.3\% & 13.74 & 0.32 \\
ExtraTrees (T+E) & 1198${}_{-60,+53}$ & 0.100 & 38.2 & 28.5 & 0.0 & 18.4\% & 191.44 & 0.76 \\
EBM (D) & 1194${}_{-60,+57}$ & 0.111 & 38.5 & 20.1 & 1.0 & 18.5\% & 7.66 & \textbf{0.05} \\
TorchMLP (T) & 1192${}_{-61,+48}$ & 0.102 & 38.6 & 30.9 & 0.0 & 17.3\% & 2832.80 & 0.11 \\
XGBoost (D) & 1187${}_{-54,+55}$ & 0.098 & 39.0 & 20.6 & 1.0 & 17.4\% & \textit{2.06} & 0.12 \\
FastaiMLP (T+E) & 1173${}_{-71,+68}$ & 0.096 & 40.0 & 30.4 & 0.0 & 18.9\% & 594.95 & 4.65 \\
ExtraTrees (T) & 1168${}_{-64,+66}$ & 0.101 & 40.3 & 26.5 & 0.0 & 19.4\% & 191.44 & 0.10 \\
RandomForest (T+E) & 1163${}_{-64,+60}$ & 0.082 & 40.7 & 32.0 & 0.0 & 19.4\% & 377.08 & 0.75 \\
LightGBM (D) & 1154${}_{-46,+48}$ & 0.071 & 41.2 & 36.2 & 0.0 & 18.1\% & 2.20 & 0.17 \\
RandomForest (T) & 1121${}_{-46,+51}$ & 0.047 & 43.5 & 37.0 & 0.0 & 20.2\% & 377.08 & 0.09 \\
FastaiMLP (T) & 1105${}_{-74,+62}$ & 0.060 & 44.5 & 32.5 & 0.0 & 20.5\% & 594.95 & 0.34 \\
PerpetualBooster (T+E) & 1084${}_{-70,+60}$ & 0.066 & 45.9 & 37.6 & 0.0 & 25.4\% & 176.26 & 0.50 \\
TabSTAR (T) & 1078${}_{-92,+82}$ & 0.109 & 46.3 & 18.7 & 0.0 & 24.4\% & 39913.02 & 4.29 \\
TabSTAR (T+E) & 1075${}_{-99,+83}$ & 0.108 & 46.5 & 16.4 & 1.0 & 24.5\% & 39913.02 & 20.17 \\
TorchMLP (D) & 1036${}_{-63,+54}$ & 0.023 & 48.8 & 44.9 & 0.0 & 22.2\% & 8.96 & 0.13 \\
PerpetualBooster (T) & 1035${}_{-66,+55}$ & 0.035 & 48.9 & 44.6 & 0.0 & 26.8\% & 176.26 & 0.19 \\
xRFM (D) & 1029${}_{-73,+73}$ & 0.049 & 49.3 & 39.2 & 0.0 & 25.0\% & 3.14 & 0.74 \\
RandomForest (D) & 1000${}_{-58,+58}$ & 0.013 & 50.9 & 44.5 & 0.0 & 25.0\% & \textit{0.43} & \textbf{0.05} \\
TabSTAR (D) & 980${}_{-115,+106}$ & 0.081 & 52.0 & 19.5 & 1.0 & 30.1\% & 398.77 & 4.65 \\
FastaiMLP (D) & 975${}_{-75,+66}$ & 0.019 & 52.2 & 48.8 & 0.0 & 24.3\% & 3.12 & 0.31 \\
KNN (T+E) & 974${}_{-82,+63}$ & 0.024 & 52.3 & 46.5 & 0.0 & 26.6\% & 129.10 & 1.63 \\
ExtraTrees (D) & 972${}_{-79,+71}$ & 0.014 & 52.4 & 48.5 & 0.0 & 26.4\% & 0.26 & \textbf{0.05} \\
PerpetualBooster (D) & 938${}_{-91,+69}$ & 0.027 & 54.1 & 42.9 & 0.0 & 31.9\% & 22.75 & 0.02 \\
Linear (T+E) & 901${}_{-98,+81}$ & 0.023 & 55.8 & 26.6 & 1.0 & 33.4\% & 240.73 & 0.31 \\
Linear (T) & 875${}_{-114,+84}$ & 0.017 & 56.9 & 36.7 & 0.0 & 33.9\% & 240.73 & 0.07 \\
KNN (T) & 824${}_{-92,+70}$ & 0.013 & 58.8 & 56.1 & 0.0 & 32.6\% & 129.10 & 0.10 \\
Linear (D) & 815${}_{-106,+84}$ & 0.005 & 59.2 & 56.4 & 0.0 & 36.4\% & 1.23 & 0.12 \\
KNN (D) & 619${}_{-109,+74}$ & 0.000 & 64.1 & 63.7 & 0.0 & 45.1\% & \textbf{0.19} & \textbf{0.04} \\
\end{longtable}
}

\section{Evaluation on the TALENT benchmark}
\label{app:talent}

We run \specmethod{} on the classification subset of the TALENT benchmark of \citet{liu_talent_2024} restricted to datasets with at most $10$ classes ($170$ binary and multiclass classification datasets). For this experiment \textbf{we use only the default configuration} for \specmethod{}. 

\textbf{Setup}:
We run three random training seeds per dataset and reuse the preprocessing pipeline of the TabArena-Lite headline run (Section~\ref{sec:exp:main}). The main baseline is the unmodified \TabICL{}v2 default, also at three seeds. The remaining baselines are the per-method, per-task numbers shipped with TALENT~\citep{liu_talent_2024}; we re-rank them in our pool rather than re-running them.

\specmethod{} completes all three seeds on every one of the $170$ in-scope datasets. Ten of those datasets, in the upper tail of sample size or feature count, required a tighter context budget than the headline default in order to fit the differentiable forward pass through \TabICL{}v2 within A100 memory and the CUDA kernel-grid limit. We cap the per-epoch in-context window at $\texttt{max\_samples}\!=\!15{,}000$ to $\texttt{60{,}000}$ rows depending on the dataset (chosen by halving from $60{,}000$ until the forward fits), and on the two largest datasets we additionally chunk the query side at $\texttt{max\_query}\!=\!8{,}192$ to keep $\texttt{train}\!+\!\texttt{query}$ tokens below the $65{,}535$ kernel-grid limit. The unmodified \TabICL{}v2 baseline reaches comparable in-context budgets via its stock sklearn predict pipeline, which subsamples internally before each forward~\citep{qu_tabiclv2_2026}, so the deviation is approximately iso-context with the baseline rather than an asymmetric advantage. 

\textbf{Result}:
Table~\ref{tab:talent-leaderboard} reports the top mean per-task rank on Accuracy over the $170$ in-scope datasets. \specmethod{} is the top-ranked method on the leaderboard at mean rank $3.83$, ahead of the \TabICL{}v2 base ($4.05$) and well clear of the next-best non-TabICL method TabR ($9.32$, $161$ datasets); every other entry on the leaderboard sits at mean rank $\geq 9.8$.

\begin{table}[t]
\caption{Mean rank on the TALENT classification subset (Accuracy, $170$ datasets with at most $10$ classes). Methods shipped with the TALENT toolbox are re-ranked over this dataset universe; their own coverage may be a subset, in which case the column $n$ reflects that subset. Best per column in \textbf{bold}.}
\label{tab:talent-leaderboard}
\centering
{\small
\begin{tabular}{lll@{ }l}
\toprule
\textbf{Metric} & \textbf{Method} & \textbf{Mean rank ($\downarrow$)} & \textbf{n} \\
\midrule
Accuracy & \textbf{\specmethod{}} & 3.83 & 170 \\
Accuracy & \TabICL{}v2 & 4.05 & 170 \\
Accuracy & TabR & 9.32 & 161 \\
Accuracy & RealMLP & 9.82 & 161 \\
Accuracy & ModernNCA & 10.10 & 159 \\
Accuracy & CatBoost & 10.42 & 161 \\
Accuracy & LightGBM & 10.78 & 161 \\
Accuracy & XGBoost & 11.97 & 161 \\
Accuracy & FTT & 12.88 & 161 \\
Accuracy & MLP\_PRL & 13.30 & 161 \\
Accuracy & Resnet & 14.94 & 161 \\
\bottomrule
\end{tabular}
}

\end{table}

\begin{table}[h]
\caption{Paired wins / ties / losses for \specmethod{} vs.\ the \TabICL{}v2 default on the TALENT classification subset restricted to datasets with at most $10$ classes ($170$ datasets per metric). ``Win'' means \specmethod{} strictly improves the seed-averaged metric over the base on that dataset.}
\label{tab:talent-h2h}
\centering
\small
\begin{tabular}{lcccc}
\toprule
\textbf{Metric}  & \textbf{Wins} & \textbf{Ties} & \textbf{Losses} \\
\midrule
Accuracy         & 84  & 14 & 72 \\
AUC              & 97  & 4  & 69 \\
Avg.\ Precision  & 84  & 11 & 75 \\
Avg.\ Recall     & 85  & 12 & 73 \\
F1               & 81  & 11 & 78 \\
LogLoss          & 109 & 0  & 61 \\
\midrule
\textbf{Total}    & \textbf{540} & \textbf{52} & \textbf{428} \\
\bottomrule
\end{tabular}
\end{table}

Head-to-head against the backbone base (\specmethod{} vs.\ \TabICL{}v2): Table~\ref{tab:talent-h2h} reports per-metric paired wins, ties, and losses across the $170$ datasets. \specmethod{} records more wins than losses on every metric. The aggregate over the $1{,}020$ paired cells is $540$ wins, $52$ ties, and $428$ losses. Restricted to the cells where the two methods produce different seed-averaged scores (i.e. Ties are excluded), the per-metric win rate is $53.8\%$ on Accuracy, $58.4\%$ on AUC, $52.8\%$ on Avg.\ Precision, $53.8\%$ on Avg.\ Recall, $50.9\%$ on F1, and $64.1\%$ on LogLoss. The F1 margin is the smallest ($81$ wins vs.\ $78$ losses, a $3$-dataset edge). The gap on the remaining metrics is more substantial, with LogLoss the most decisive.




\end{document}